\documentclass{article}

\PassOptionsToPackage{numbers, compress, square}{natbib}

\usepackage[preprint]{neurips_2023}

\usepackage{subcaption}
\usepackage{xr}
\usepackage[utf8]{inputenc} 
\usepackage[T1]{fontenc}    
\usepackage{url}            
\usepackage{booktabs}       
\usepackage{amsfonts}       
\usepackage{nicefrac}       
\usepackage{microtype}      
\usepackage{xcolor}         

\usepackage{amsmath}
\usepackage{amssymb}
\usepackage{bbm}
\usepackage{graphicx,psfrag,epsf}
\usepackage{enumerate}
\usepackage{color}
\usepackage{algorithm,algorithmicx,algpseudocode}
\usepackage{url}
\usepackage{enumitem}
\usepackage{array,booktabs}
\usepackage{amsthm}

\newcommand{\Cov}{\operatorname{Cov}}

\newcommand{\logit}{\operatorname{logit}}
\newcommand{\Var}{\operatorname{Var}}
\newcommand{\sign}{\operatorname{sign}}

\newtheorem{theorem}{Theorem}

\newtheorem{assumption}{Assumption}

\title{Is this model reliable for everyone?\\Testing for strong calibration}

\author{%
Jean Feng\thanks{Department of Epidemiology and Biostatistics, University of California, San Francisco} \\
 \And
Alexej Gossmann\thanks{Center  for  Devices  and  Radiological  Health, U.S. Food and Drug Administration}\\
\And
Romain Pirracchio\thanks{Department of Anesthesiology, University of California, San Francisco}\\
\And
Nicholas Petrick$^\dagger$\\
\And
Gene Pennello$^\dagger$\\
\And
Berkman Sahiner$^\dagger$
}

\begin{document}

\maketitle

\begin{abstract}
In a well-calibrated risk prediction model, the average predicted probability is close to the true event rate for any given subgroup. Such models are reliable across heterogeneous populations and satisfy strong notions of algorithmic fairness. However, the task of auditing a model for strong calibration is well-known to be difficult---particularly for machine learning (ML) algorithms---due to the sheer number of potential subgroups. As such, common practice is to only assess calibration with respect to a few predefined subgroups. Recent developments in goodness-of-fit testing offer potential solutions but are not designed for settings with weak signal or where the poorly calibrated subgroup is small, as they either overly subdivide the data or fail to divide the data at all. We introduce a new testing procedure  based on the following insight: if we can reorder observations by their expected residuals, there should be a change in the association between the predicted and observed residuals along this sequence if a poorly calibrated subgroup exists. This lets us reframe the problem of calibration testing into one of changepoint detection, for which powerful methods already exist. We begin with introducing a sample-splitting procedure where a portion of the data is used to train a suite of candidate models for predicting the residual, and the remaining data are used to perform a score-based cumulative sum (CUSUM) test. To further improve power, we then extend this adaptive CUSUM test to incorporate cross-validation, while maintaining Type I error control under minimal assumptions. Compared to existing methods, the proposed procedure consistently achieved higher power in simulation studies and more than doubled the power when auditing a mortality risk prediction model.
\end{abstract}

\section{Introduction}
Calibration is a fundamental measure of model reliability: a risk prediction model is calibrated for a subgroup if the average predicted probability corresponds to the observed event rate.
When decisions are made using absolute risk thresholds---as is common in medicine \citep{Goff}---calibration has also been shown to directly impact the utility of a model \citep{Van_Calster2015-ob}.
However, machine learning (ML) algorithms are typically trained to optimize average performance and can be poorly calibrated in particular subgroups \citep{Chatterjee2016-lb,Barda2021-qu}, leading to concerns regarding their robustness and fairness.
In fact, performance can be particularly low for subgroups defined by interactions of multiple variables (e.g. race and gender), an issue known as intersectionality \citep{Buolamwini2018-cb}.
Ideally, a risk prediction model is ``strongly'' calibrated, in that it is calibrated for all individuals or, equivalently, all possible subgroups \citep{Van_Calster2016-ey, Zhao2020-md}.

Unfortunately, achieving or even verifying strong calibration is challenging due to the curse of dimensionality: as the number of variables grows, the subgroups---and thus the number of observations per subgroup---get smaller.
As such, much of this research has been focused on ``moderate'' calibration, which only require calibration with respect to a small set of predefined subgroups \citep{Brown1975-sx, Tsiatis1980-fp, Hawkins1991-cx, Hosmer1997-ss, Hosmer2002-mp, Lin2002-gu, DiCiccio2020-yo}.
These methods are widely used among practitioners, given their ease of use and applicability to small datasets.
With the recent interest in algorithmic fairness and model reliability, many recent works have sought to achieve stronger forms of calibration, either by identifying subgroups that require revision \citep{Chung2019-tz, Eyuboglu2022-ho} or revising the model directly \citep{Hebert-Johnson2018-vb, Kim2019-bo, Luo2022-tg}.
However, these methods are either meant to be exploratory or only provide statistical guarantees when the number of observations is sufficiently large.
The minimum sample size is typically at least tens or even hundreds of thousands of observations, which is often unrealistic in many settings.

Rather than tackling the difficult task of subgroup identification or model revision, we instead consider the problem of testing if there \textit{exists} any poorly calibrated subgroup.
This is much more feasible in settings with limited data or lower signal-to-noise ratios, and still answers the important yes/no question ``Is this ML algorithm reliable for everyone?''
Moreover, the answer to this question can help decide if more sophisticated data-hungry procedures are necessary.

We formalize this as the following hypothesis test.
Let $\hat{p}:\mathcal{X} \mapsto [0,1] $ be the risk prediction algorithm and $p_0:\mathcal{X} \mapsto [0,1] $ be the true 
 event rate over some domain $\mathcal{X} \in \mathbb{R}^d$, where $d$ is the number of variables.
For some pre-specified tolerance level $\delta$, define the poorly calibrated subgroup as
$
A_{\delta} = \left\{
    x \in \mathcal{X}: \left|\hat{p}(x)  - p_0(x) \right| > \delta
    \right\}.
$
The hypothesis test checks if the set $A_{\delta}$ is too large, i.e.
\begin{align}
    \begin{split}
        H_0 &: \Pr\left(X \in A_{\delta} \right) \le \epsilon \\
        H_1 &: \Pr\left(X \in A_{\delta} \right) > \epsilon
    \end{split}
    \label{eq:hypo_test}
\end{align}
from some minimally acceptable prevalence $\epsilon \ge 0$.
\citep{Hudson2021-gm} considers a similar hypothesis test, but only for the univariate case ($d=1$) and where $\delta = 0$.
Unfortunately, it is unclear how to extend their proposal to settings with larger $d$ or $\delta > 0$.

In settings with larger $d$, the most relevant works are \citep{Jankova2020-kl} and \citep{Zhang2021-oh} from the goodness-of-fit (GOF) testing literature.
Although GOF tests technically answer a different question from \eqref{eq:hypo_test}, it is relatively straightforward to extend these methods to test for strong calibration.
To address the curse of dimensionality, both procedures perform sample-splitting: a portion of the data is used to train a random forest (RF) to predict the residuals and the remaining data are used to test for GOF with respect to the learned residual model.
The difference between \citep{Jankova2020-kl} and \citep{Zhang2021-oh} is primarily in the second step.
The former assesses the association between the predicted and observed residuals with respect to the entire population through a score test, so it fails to reflect subpopulation structure.
The latter bins observations with similar predicted residuals and performs a Chi-squared test, but no information is borrowed across bins and the procedure can be highly sensitive to the number of bins.
As such, there are a number of limitations with these works.
First, these methods are not well-suited for settings where only a subgroup is poorly calibrated, particularly when this subgroup is small.
In addition, both papers only use RFs to model the residuals, but we show that RFs are weak at extracting the remaining signal after fitting a tree-based risk prediction model.
Instead, it is important to consider a diverse pool of candidate residual models.
Also, there are no extensions of these methods that use cross-validation (CV) while maintaining Type I error control.
Finally, to fit the residual model, both procedures perform further sample splitting to tune the hyperparameters.
This can be noisy in small sample sizes and lead to an suboptimal choice of hyperparameters.
It is also computationally expensive when combined with CV.

We introduce a more powerful testing procedure for strong calibration motivated with the following insight: if observations are ordered by their predicted residuals, we expect the association between the observed and predicted residuals to drop somewhere along this sequence if a poorly calibrated subgroup exists.
Our key contributions are (i) we show how reframing the problem of detecting a poorly calibrated subgroup into that of changepoint detection substantially improves power because the changepoint structure closely mimics the true subpopulation structure; (ii) we demonstrate how additional gains in power and computational efficiency can be made by fitting a pool of candidate residual models and performing a suite of structural change tests; (iii)
we incorporate CV into the procedure to further improve power, while maintaining Type I error control under much weaker assumptions than prior works; and (iv) we provide visualization tools to aid model diagnosis.
In experiments, the proposed procedure significantly outperforms existing methods.
Code is available at \url{https://github.com/jjfeng/testing_strong_calibration}.

\subsection{Related works}
\textbf{Individual/metric fairness.}
Strong calibration of a ML algorithm is only one approach to measuring model fairness.
Other common measures of algorithmic fairness are concerned with statistical parity or balance of error rates between subgroups \citep{Hardt2016-fc, Mitchell2021-wq}.
Similar to the critiques of moderate calibration, group-wise equality in error rates has been critized for being too coarse \citep{Dwork2012-aj}.
Recent works have aimed for individual or metric fairness to ensure similar performance between similar individuals \citep{Ilvento2020-bs,Ruoss2020-no}, and hypothesis tests for evaluating individual fairness have been developed \citep{Xue2020-nt, Maity2021-nm}.
However, unlike the proposed procedure, these methods assume a similarity metric is known a priori, which corresponds to prespecifying the subgroup structure.

\textbf{Conformal inference.}
Recent works have highlighted how the coverage rate guarantees from conformal inference procedures can be used to calibrate risk prediction algorithms \citep{Vovk2020-pi, Marx2022-jh}.
Ordinary conformal inference procedures only guarantee marginal coverage rates \citep{Vovk2005-wf}, which satisfy notions of weak calibration \citep{Van_Calster2016-ey}.
More recent works have extended these methods to provide guarantees with respect to predefined subgroups \citep{Vovk2013-ou,Lei2014-uh,Romano2020-ua} and weighted neighborhoods \citep{Guan2023-is}.
Taking such guarantees to the limit, \citep{Foygel_Barber2021-gx} proved that it is impossible for a non-trivial procedure to guarantee uniform conditional coverage rates.
Our ability to test for strong calibration does not contradict this impossibility result and provides instead a complementary (and perhaps more positive) result.
Because hypothesis tests start from the opposite angle of ``innocent until proven guilty,'' we can at least determine if there is sufficient evidence that a given model fails to satisfy strong calibration.

\textbf{Distributionally robust optimization (DRO).}
Much of DRO is concerned with training models that minimize the worst-case performance over some set of distributional perturbations \citep{Ben-Tal, Duchi2021-un, Duchi2022-kt}.
Based on these ideas, recent works propose to estimate the worst-case performance of a given ML algorithm over all subgroups with size $\epsilon >0$ \citep{Subbaswamy2021-np, Li2021-bk}.
\citep{Subbaswamy2021-np} is closest in spirit to this work, in that it performs statistical inference to output confidence intervals (CI).
In fact, it similarly splits the data to estimate a model of the ML algorithm's error on one partition and assess worst-case performance on the other partition.
Nevertheless, the method differs in that it requires much larger sample sizes, $\epsilon$ to be bounded away from zero, and the error model to converge at a fast enough rate.
In contrast, our proposed procedure is suitable for smaller sample sizes, can test for arbitrarily small subgroups, provides \textit{assumption-free, finite-sample} Type I error control when performing a single sample-split, and requires much weaker assumptions to establish asymptotic Type I error control when using CV.

\section{Method}
\label{sec:method}

For ease of exposition, we begin with the one-sided testing problem where we replace $A_{\delta}$ with the one-sided violation set $A_{\delta,>} = \{x \in \mathcal{X} : p_0(x) - \hat{p}(x) > \delta \}$.
The first step is to reformulate the hypothesis test as a score test.
In the main text of this paper, we focus on the test where $\epsilon = 0$.
In the Appendix, we describe how the proposed procedure can be easily extended to address non-zero $\epsilon$.

Let $\mathcal{H}_+$ be the class of bounded non-negative real-valued functions.
For a given $h \in \mathcal{H}_+$, define a working model for the structural change of the log odds (logit) to be
\begin{align}
\logit\left(p(X; h)\right)
= \logit(\hat{p}_\delta(X)) + \theta h(X),
\label{eq:working_model}
\end{align}
where $\hat{p}_{\delta}(X) = \left[\hat{p}(X) + \delta\right]_{[0,1]}$ and $q\mapsto [q]_{[0,1]}$ is a projection into the range of valid probabilities $[0,1]$.
The gradient of the log likelihood, also known as the score, at $\theta = 0$ is equal to
\begin{align}
\dot{\ell}(Y|X;h)
=
\left.\frac{\partial}{\partial \theta} \log p(Y|X; h) \right|_{\theta=0}
=
\left(Y - \hat{p}_{\delta}(X)  \right)h(X).
\label{eq:score_def}
\end{align}
In the set $A_{\delta,>}$, the expected score $\mathbb{E}[\dot{\ell}(Y|X;h)|X]$ is positive if $h(X)$ is positive.
Outside of this set, the expected score is non-positive.
As such, we will refer to $h$ as a detector.
We can rewrite the one-sided hypothesis test in terms of the maximum expected score over detectors in $\mathcal{H}_+$, i.e.
\begin{align}
\begin{split}
H_{0,>} &:
\sup_{h \in \mathcal{H}_+}
\mathbb{E}\left[\left(Y - \hat{p}_{\delta}(X)  \right)h(X) \right]
\le 0\\
H_{1,>} &:
\sup_{h \in \mathcal{H}_+}
\mathbb{E}\left[\left(Y - \hat{p}_{\delta}(X)  \right)h(X) \right]
> 0.
\label{eq:one_side_score}
\end{split}
\end{align}

In practice, it is computationally infeasible to test the entire set $\mathcal{H}_+$.
Instead, we will generate a subset $\widehat{\mathcal{H}}_+ \subseteq \mathcal{H}_+$ to replace $\mathcal{H}_+$ in \eqref{eq:one_side_score}, resulting in a \textit{restricted} score test.
Thus a procedure with Type I error control for a restricted score test also satisfies Type I error control for \eqref{eq:one_side_score}.

In the following sections, we introduce the testing procedure using single sample-split, extend it to incorporate CV, and finally extend it to the two-sided setting.

\begin{figure}
    \centering
    \vspace{-0.1in}
    \includegraphics[width=\textwidth]{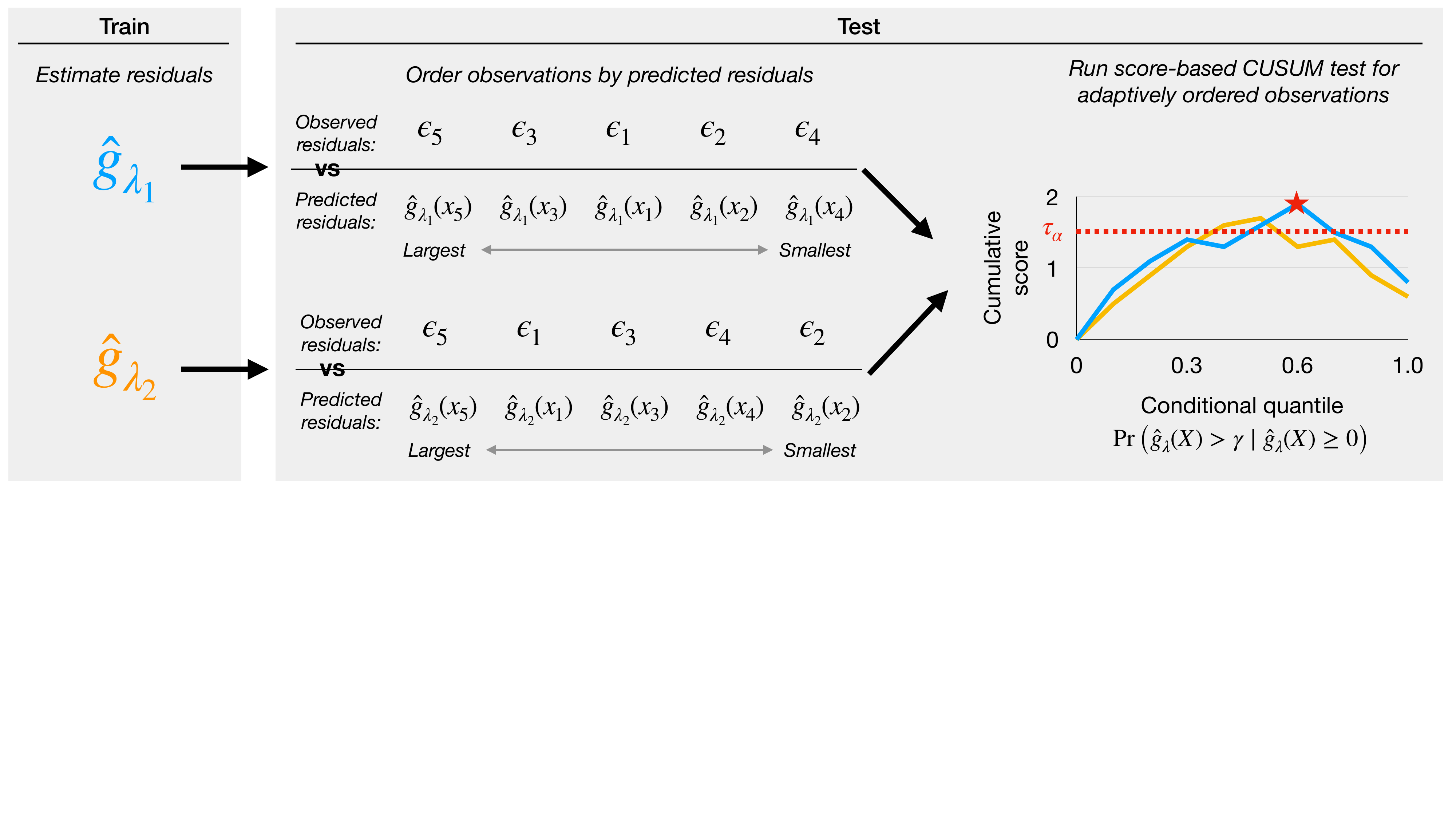}
    \vspace{-1.5in}
    \caption{
    Summary of the procedure for testing if there exists a subgroup for whom the predicted probabilities are too small.
    After an ensemble of models are trained to predict the residual on one portion of the data, we order the remaining observations by their predicted residuals and run changepoint tests.
    Control charts plot the cumulative score and the the test statistic (red star) is the maximum value attained across all curves.
    The test is rejected if the test statistic exceeds the critical value $\tau_{\alpha}$.
    Extension to the two-sided setting orders observations by the absolute predicted residuals.
    }
    \vspace{-0.2in}
    \label{fig:test_outline}
\end{figure}

\subsection{Sample-splitting}
\label{sec:method_split}

Suppose the audit data are composed of independent and identically distributed (IID) observations with variables $X_i \in \mathcal{X}$ and binary outcome $Y_i$ for $i = 1,\cdots, n$.
The outline for the sample-splitting procedure is as follows.
Let the first $n_1$ observations form a training partition and the remaining $n_2 = n - n_1$ observations form a test partition.
Using the training data, we generate a set of candidate detectors $\widehat{\mathcal{H}}_{+,\Lambda}$ across different hyperparameter settings $\Lambda$.
Using the test data, we calculate the maximum empirical score over the set of candidate detectors as our test statistic, i.e.
\begin{align}
\hat{T}_{n,>}^{(split)} = \sup_{h \in \widehat{\mathcal{H}}_{+,\Lambda}}
\frac{1}{n_2}
\sum_{i=n_1 + 1}^{n}
(Y_i - \hat{p}_{\delta}(X_i)) h(X_i).
\label{eq:sample_split_test_stat}
\end{align}
We reject the null hypothesis if $\hat{T}_{n,>}^{(split)}$ exceeds the critical value $\tau_{\alpha}$ defined in the theorem below.

As described below, we use a simple Monte Carlo procedure to calculate the critical value which, unlike existing tests for model calibration, provides \textit{finite-sample} Type I error control.
In particular, using a coupling argument, we prove that the distribution of the test statistic is stochastically largest at the boundary of the null hypothesis space where $p_0(x) = \hat{p}_\delta(x)$.
Proofs for all the theoretical results are in the Appendix.
\begin{theorem}
    Let $Y^*_i$ be the binary random variable with probability $\hat{p}_{\delta}(X_i)$.
    By setting $\tau_{\alpha}$ to be the $1-\alpha$ quantile of 
    \begin{align}
        T_{>}^{*(split)} = \sup_{h \in \widehat{\mathcal{H}}_{+,\Lambda}}
        \frac{1}{n_2}
        \sum_{i=n_1 + 1}^{n}
        (Y^*_i - \hat{p}_{\delta}(X_i)) h(X_i),
    \end{align}
    the Type I error of the sample-splitting test is controlled at level $\alpha$.
\label{theorem:super_unif}
\end{theorem}

Given the Type I error guarantee, the key question is how to construct a set of detectors that maximizes power.
As motivation, suppose we were only allowed to generate a single detector $h$.
Per \citep{vaart_1998}, the local asymptotic power of the test with respect to $h$ is determined by the ratio
\begin{align}
\frac{\mathbb{E}\left[(Y - \hat{p}_{\delta}(X))h(X)\right]}{\sqrt{\Var\left((Y - \hat{p}_{\delta}(X))h(X) \right)}}.
\label{eq:slope}
\end{align}
Let $g_0(X) = p_0(X) - \hat{p}_{\delta}(X)$ denote the expected residuals.
Given the constraint that detectors must be non-negative, the numerator is maximized by $g_0(X)\mathbbm{1}\{g_0(X) \ge 0\}$.
To maximize the ratio, we can tune over the broader class of detectors $h_{0,\gamma}(X) = g_0(X) \mathbbm{1}\{g_0(X) > \gamma\}$ for $\gamma \ge 0$, which also reflects our interest in observations with the largest values of $g_0(X)$.
This family of detectors has another advantage in practice, where $g_0$ is unknown and must be estimated.
Generating detectors based on an estimate of the expected residuals, the expected score is large as long as thresholding on the predicted residuals isolates a subset of observations with large $g_0(X)$, even if the predicted residuals are not accurate for the entire population.

Given this motivation, we propose the following procedure for generating detectors.
Suppose one has a set of candidate algorithms (e.g. random forests and neural networks) for estimating the residuals, indexed by the set of hyperparameters $\Lambda$.
We fit residual models $\hat{g}_{\lambda,n}$  for each $\lambda \in \Lambda$ on the training partition, and construct the set of detectors
\begin{align}
\hat{\mathcal{H}}_{+,\Lambda} = \left\{
\hat{h}_{\lambda, \gamma,n} =
\hat{g}_{\lambda,n}(X) \mathbbm{1}\{\hat{g}_{\lambda,n}(X) > \gamma\}: \gamma  \ge 0, \lambda \in \Lambda
\right \}.
\label{eq:detectors}
\end{align}
The test statistic can now be rewritten as
\begin{align}
    \hat{T}_{n,>}^{(split)} =
    \max_{\lambda \in \Lambda}
    \underbrace{
    \max_{\gamma \ge 0}
    \sum_{i=n_1 + 1}^{n}
    \left(Y_{i} - \hat{p}_{\delta}(X_{i})\right)
    \hat{g}_{\lambda,n}(X_{i})
     \mathbbm{1}\left\{
        \hat{g}_{\lambda, n}(X_i) \ge \gamma
        \right\}
    }_{\text{Score-based CUSUM}}.
\end{align}
Notice that the inner summation corresponds exactly to the score-based cumulative sum (CUSUM) test statistic, which is typically used to detect changepoints along a single axis \citep{Gombay2003-ln, Gombay2017-wb, Feng2022-az}.
So our procedure can be viewed as performing changepoint detection along data-adaptively defined axes $\{\hat{g}_{\lambda,n}: \lambda \in \Lambda\}$, where the null hypothesis of each changepoint test is that the mean score is uniformly non-negative and the alternative is that there is some point after which the mean score is positive.

Leveraging this connection with the changepoint literature, we can visualize the test using ``control charts,'' which are typically used to visualize changepoint detection procedures along a single dimension \citep{Montgomery2013-sk, Feng2022-mk}.
As shown in the example in Figure~\ref{fig:test_outline}, each curve corresponds to the cumulative score when observations are ordered by their predicted residuals, and the maximum value attained across all curves corresponds to the test statistic.
Large positive slopes correspond to subgroups where the model is very poorly calibrated, whereas flat or negative slopes correspond to subgroups where model calibration is mostly within the desired tolerance.
The location of the peak indicates the size of the poorly calibrated subgroup detected by the procedure.
Thus the shape of the curve provides insight into the nature of the poorly calibrated subgroup.

Finally, one may ask (i) how many candidate residual models should one fit and (ii) how much data should one dedicate to training the residual models?
These questions concern two tradeoffs.
More residual models increase our chance of finding a changepoint but require more stringent multiplicity correction.
Also, allocating more data to training can increase the accuracy of the residual models but reduces the sample size available for testing.
We clarify the answer to these two questions in the following result.
Note that $c_k$ denote positive constants below.
\begin{theorem}
        Consider any $\gamma \ge 0, \lambda \in \Lambda,$ and $\omega \le 1$.
        Suppose $n_1$ is chosen so that
	$
	\mathbb{E}\left[\left(Y - \hat{p}_{\delta}(X)\right) h_{0,\gamma}(X) \right]
	- c_1 n_1^{-\omega/2}
	\ge c_2 \sqrt{\frac{\log(|\Lambda| (n_2 + 1)/\alpha)}{n_2}}.
	$
        Conditional on the event that
	\begin{align}
	\mathcal{S}^{(n_1)}_{\lambda,\omega} &= \left\{
        \mathbb{E} \left\|h_{0,\gamma}(X)- \hat{h}_{\lambda,\gamma,n}(X)\right\|^2 \le c_3 n_1^{-\omega}
	\right\},
	\label{eq:convergence}
	\end{align}
	statistical power is lower bounded by
	\begin{align}
 \hspace{-0.4in}
	\Pr\left(\hat{T}_{n,>}^{(split)} > \tau_{\alpha} \mid \mathcal{S}^{(n_1)}_{\lambda,\omega} \right )
        \ge 1 - \exp\left(
	-\frac{
		n_2\left(\mathbb{E}\left[\left(Y - \hat{p}_{\delta}(X)\right) h_{0,\gamma}(X) \right]
		- c_1 n_1^{-\omega/2}
		- c_4 \sqrt{\log (|\Lambda|(n_2 + 1)/\alpha)/n_2}
		\right)^2
	}{
		2 c_5
	}
	\right),
	\label{eq:power_train_test}
	\end{align}
        where $|\Lambda|$ is the number of hyperparameters.
	\label{theorem:power}
\end{theorem}
\noindent 
The lower bound in \eqref{eq:power_train_test} states that the number of hyperparameters impacts power only through a logarithmic term, which justifies our approach of testing a suite of residual models.
In contrast, existing methods test only a single residual model \citep{Jankova2020-kl, Zhang2021-oh}, which is tuned using CV within the training partition.
In settings with limited amounts of audit data, CV tends to overfit and select a suboptimal detector, leading to lower power.
Moreover, this procedure is very computationally expensive when combined with CV, because one would have to perform CV within CV.

Second, the numerator in \eqref{eq:power_train_test} grows linearly with the amount of test data and slowly with the amount of training data.
This highlights an interesting ``phase change'' in how much data one should allocate to training versus testing.
One should allocate just enough training data so that \eqref{eq:convergence} is satisfied with high probability (note that many ML algorithms have convergence rates of this form) and dedicate the rest to testing.

\subsection{$K$-fold Cross-validation}
We now extend the sample-splitting procedure to use CV to further improve power.
The technical challenge is how to maintain Type I error control, despite the correlation between estimators across folds.
Prior works provide only ad-hoc solutions \citep{Zhang2021-oh} or assume estimators converge to the oracle sufficiently fast \cite{Subbaswamy2021-np}.
Here we present a procedure that requires very minimal assumptions.

We extend the sample-splitting procedure as follows.
Partition the audit data into folds $V_k$ for $k = 1,\ldots, K$.
For each $\lambda\in \Lambda$, let $\hat{g}_{\lambda, n}^{(-k)}$ denote the estimated residual model using data in all but the $k$-th fold for $k=1,\ldots,K$.
The CV test statistic is defined as
\begin{align}
\hat{T}_{n,>}^{(CV)} =
\sup_{\lambda \in \Lambda, \gamma \ge 0}
\sum_{k=1}^K
\sum_{(X_i, Y_i)\in V_k}
\left(Y_i - \hat{p}_{\delta}(X_i)\right)
\hat{g}_{\lambda, n}^{(-k)}(X_i) \mathbbm{1}\left\{
\hat{g}_{\lambda, n}^{(-k)}(X_i) \ge \gamma
\right\}.
\label{eq:cv_stat}
\end{align}

To establish Type I error control, we only require the following uniform convergence assumption.
\begin{assumption}
    For a given $\lambda$ and $\gamma$, define $\bar{h}_{\lambda,\gamma,n}$ as the average detector estimated from $K-1$ folds of a dataset with $n$ observations.
    That is, 
	$
	\bar{h}_{\lambda,\gamma,{n}} = \mathbb{E}\left[
        \hat{g}_{\lambda,{n}}^{(-1)}(X) \mathbbm{1}\left\{
        \hat{g}_{\lambda,{n}}^{(-1)}(X) \ge \gamma
        \right\}
	\right]$,
	where the expectation is with respect to the estimated residual models.
	Suppose that
	\begin{align}
	\sup_{\lambda \in \Lambda, \gamma \ge 0}
	\left \|
	\bar{h}_{\lambda,\gamma,{n}}
	- \hat{h}_{\lambda,\gamma,{n}}^{(-1)}
	\right \|_{2} \rightarrow_p 0.
	\label{eq:g_conv}
	\end{align}
 \label{assump:converges}
\end{assumption}
\vspace{-0.1in}

Under this assumption, we prove that one can essentially treat the estimated residual models as fixed and use the same Monte Carlo procedure as before to calculate the critical value.
The only difference is that we control the Type I error rate asymptotically, rather than in finite samples.
\begin{theorem}
 Suppose Assumption~\ref{assump:converges} holds.
    Define $Y^*_i$ using the definition in Theorem~\ref{theorem:super_unif} and $T_{>}^{*(CV)}$ using \eqref{eq:cv_stat} but replace $Y_i$ with $Y^*_i$.
    If $\tau_{\alpha}$ is set to the $1-\alpha$ quantile of $T_{>}^{*(CV)}$,
    the Type I error of the CV test is asymptotically controlled at level $\alpha$.
	\label{theorem:cv}
\end{theorem}

\subsection{Extension to the two-sided test}
\label{sec:method_two}
Finally, we extend the procedure to test the two-sided null hypothesis.
The key difference is that rather than ordering observations by their predicted residuals, we now order observations by the magnitude of the predicted residuals.
For ease of exposition, we only describe the sample-splitting procedure for the two-sided setting. The same ideas are used to extend the CV procedure.

Following the same logic as before, we begin with restating the two-sided hypothesis test in terms of a score test.
Let $\mathcal{H}$ refer to the set of bounded functions, removing the prior restriction of non-negativity.
For a given $h$, the working model for structural change is now
$
\logit(p(X;h)) = \logit(\hat{p}_{\delta \sign(h)}) + \theta h(X),
$
where $\sign(h(X))$ is the sign of $h(X)$ and zero if $h(X) = 0$, and $\hat{p}_{\delta \sign(h)}(X) = \left[\hat{p}(X) + \delta \sign(h(X))\right]_{[0,1]}$.
Thus we can reframe \eqref{eq:hypo_test} as
\begin{align}
\begin{split}
H_0 &:
\sup_{h \in \mathcal{H}}
\mathbb{E}\left[\left(Y - \hat{p}_{\delta \sign(h) } \right)h(X) \right]
\le 0\\
H_1 &:
\sup_{h \in \mathcal{H}}
\mathbb{E}\left[\left(Y - \hat{p}_{\delta \sign(h)}  \right)h(X) \right]
> 0.
\label{eq:two_side_score}
\end{split}
\end{align}

As before, we will instead perform a restricted score test by generating candidate detectors given a set of hyperparameters $\Lambda$.
More specifically, for each $\lambda\in \Lambda$, we fit residual models $\hat{g}_{\lambda,n}(X)$ that estimate $p_0(X) - \hat{p}_{\delta}(X)$ if $p_0(X)> \hat{p}_{\delta}(X)$, $p_0(X) - \hat{p}_{-\delta}(X)$ if $p_0(X)< \hat{p}_{-\delta}(X)$, and zero otherwise.
We then generate detectors
$
\hat{h}_{\lambda,\gamma,n}(x) = \hat{g}_{\lambda,n}(X) \mathbbm{1}\left\{|\hat{g}_{\lambda,n}(X)| \ge \gamma \right\}
$ for $\gamma \ge 0$.
Consequently, the test statistic in the two-sided setting is the maximum of the score-based CUSUM statistics where observations are ordered by the absolute predicted residuals, i.e.
\begin{align}
    \hat{T}^{(split)}_{n}= 
    \max_{\lambda \in \Lambda, \gamma \ge 0}
    \sum_{i=n_1+1}^n 
    \left(Y_{i} - \hat{p}_{\delta \sign(h_{\lambda, \gamma,n})}(X_{i})\right)
    \hat{g}_{\lambda,n}(X_{i})
     \mathbbm{1}\left\{
        \left|\hat{g}_{\lambda, n}(X_i) \right| \ge \gamma
        \right\}.
    \label{eq:test_stat_2side}
\end{align}

To calculate the critical value, we must modify the Monte Carlo procedure.
Unlike the one-sided setting, it is no longer straightforward to determine the null distribution whose test statistic is stochastically largest, because estimated models may disagree on the sign of the expected residual.
As such, we set the critical value to the quantile of a \textit{modified} statistic that upper bounds \eqref{eq:test_stat_2side}.
More specifically, for each $X$, we sample \textit{two} binary outcomes with marginal probabilities $\hat{p}_{\delta}(X)$ and $\hat{p}_{-\delta}(X)$.
For each model $\hat{g}_{\lambda,n}$, we calculate a ``bounding'' CUSUM statistic by selecting the outcome generated with probability $\hat{p}_{\delta}(X)$ if the predicted residual is positive and $\hat{p}_{-\delta}(X)$ if the predicted residual is negative.
The modified statistic is the maximum of these bounding CUSUM statistics.
This procedure, formally described below, ensures finite-sample Type I error control.
\begin{theorem}
	Let $U_i$ for $i = 1,\ldots, n$ be IID standard uniform random variables.
	Define
        $
	Y_{i,\lambda}^* = \mathbbm{1}\left\{
	U_i \le \hat{p}(X_i) + \delta \sign(\hat{g}_{\lambda,n}(X_i))
	\right\}.
	$
	Define ${T}^{* (split)}$ using \eqref{eq:test_stat_2side} but replacing $Y_i$ with $Y_{i,\lambda}^*$.
        Set the critical value $\tau_{\alpha}$ to the $1-\alpha$ quantile of $T^{*(split)}$, 
	For the two-sided hypothesis test, the sample-splitting procedure that rejects the null when $\hat{T}_n^{(split)} > \tau_{\alpha}$ controls the Type I error at level $\alpha$.
	\label{theorem:super_unif_two_side}
\end{theorem}

\vspace{-0.2in}
\subsection{Variable importance plots}

In addition to control charts, we can use variable importance (VI) plots to gain insight into potential reasons for model miscalibration.
Here we consider a simple procedure using permutation VI to compute how important each variable is for detecting a poorly calibrated subgroup.
(Future work may consider more sophisticated VI measures such as using Shapley values \citep{Lundberg2017-xl, Williamson2020-lx}.)
For each variable, we permute its values and calculate the change in the test statistic.
The importance of that variable is defined as the drop in the test statistic, where a larger drop indicates a more important variable.
We emphasize that this definition of VI is \textit{not} the same as ordinary VI measures that quantify how useful a variable is to a model's average performance.
Ordinary VI measures describe the majority group and are not meant to characterize poorly calibrated subgroups.



\vspace{-0.1in}
\section{Simulations}
\label{sec:simulations}

\begin{table}
    \centering
    \vspace{-0.1in}
    \hspace{-0.3in}
      \begin{tabular}{p{1.2in}|p{1.9in}|p{2.4in}}
      & Chi-squared tests & Score-based tests \\
      \hline
      Prespecified axis & $\bullet$ Hosmer-Lemeshow: \texttt{ChiSq}  & $\bullet$ Score test for Platt scaling: \texttt{Score} \\
      \hline
     Data-adaptive axes&
     $\bullet$ Based on \citep{Zhang2021-oh}: \texttt{AdaptiveChiSq} & $\bullet$ Based on \citep{Jankova2020-kl}: \texttt{AdaptiveScoreSimple} 
     \\
     & & $\bullet$ Proposed: \texttt{AdaptiveScoreCUSUM}
	\end{tabular}
	\caption{Categorization of existing and proposed testing procedures}
	\label{table:comparators}
 \vspace{-0.2in}
\end{table}

We begin with a simulation study comparing the proposed procedure (\texttt{AdaptiveScoreCUSUM}) to four procedures: the Hosmer-Lemeshow test (\texttt{ChiSq}) \citep{Lemeshow2013-mk}, a score test based on Platt scaling (\texttt{Score}) \citep{Platt1999-lm}, an adaptive Chi-squared test that extends \citep{Zhang2021-oh} (\texttt{AdaptiveChiSq}), and an adaptive score test that extends \citep{Jankova2020-kl} (\texttt{AdaptiveScoreSimple}).
The two comparator score tests can be viewed as special cases of our procedure.
\texttt{Score} implements a score-based CUSUM test with respect to the logistic recalibration model  \eqref{eq:working_model} along the prespecified axis $\logit(\hat{p}(X))$.
This test is not data-adaptive, so it does not require sample-splitting.
\texttt{AdaptiveScoreSimple} is implemented to be almost the same as ours except, following \citep{Jankova2020-kl}, we do not explicitly create subgroup detectors and fix $\gamma = 0$.
The procedures can be categorized with respect to the 2x2 grid shown in Table~\ref{table:comparators}, where the rows correspond to test procedures that prespecify versus adaptively define axes, and columns correspond to procedures that run Chi-squared versus score tests.
For tests that prespecify axes, we follow standard practice and use the single axis $\hat{p}(X)$.
For all methods, we use the proposed Monte Carlo procedure to calculate critical values and control Type I error.

For the Chi-squared tests, we present results for when the data are divided into 2 versus 10 bins.
We had also tested bin numbers in between, but the performance was similar or worse.
Because Chi-squared tests are traditionally designed to test hypotheses with a tolerance of $\delta = 0$, we modified the test statistic to test non-zero tolerances.

The data-adaptive tests were implemented using 4-fold CV and made to be as comparable as possible.
To this end, we extend and unify tests in \citep{Jankova2020-kl} and \citep{Zhang2021-oh} using our framework.
We fit residual models using RFs and kernel logistic regression across various hyperparameter settings.
The detectors use as input variables $x_1,\ldots,x_{d}$ as well as the predicted logit from the risk prediction algorithm.

We simulate data as follows.
Covariates $X \in \mathbb{R}^{10}$ were independently sampled from Uniform$[-5,5]$.
The outcome is sampled with the log odds as
\begin{align}
(0.6x_0 + 0.4x_2 + 0.2x_3) \mathbbm{1}\{\max(x_1, -x_2) \ge -2\}
+ 0.2x_1\mathbbm{1}\{\max(x_1, -x_2) < -2 \}.
\label{eq:data_sim}
\end{align}
The Appendix includes additional simulation details and results as well as a second simulation study verifying Type I error control of the proposed procedure.

\begin{figure}
	\centering
 \vspace{-0.1in}
 \hspace{-0.8in}
     \includegraphics[width=0.8\textwidth]{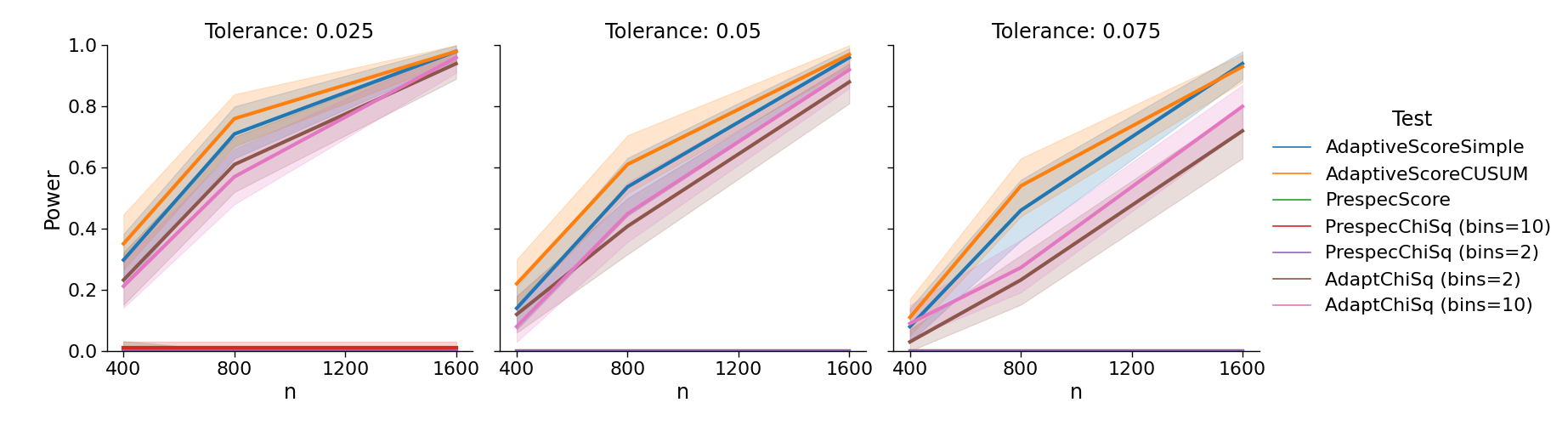}
	\includegraphics[width=0.28\textwidth]{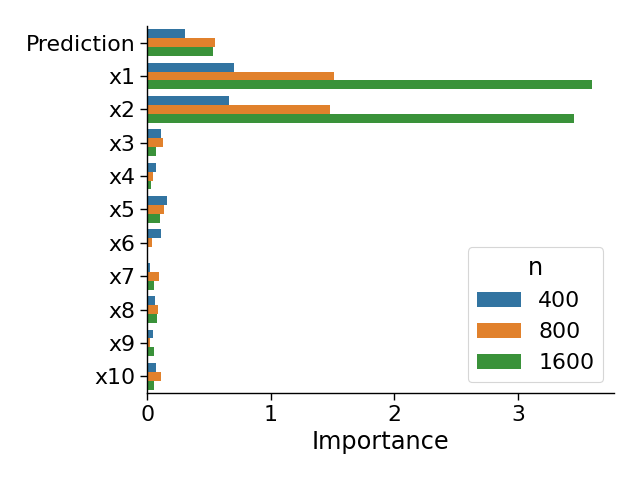}
 
\hspace{-0.8in}
     \includegraphics[width=0.8\textwidth]{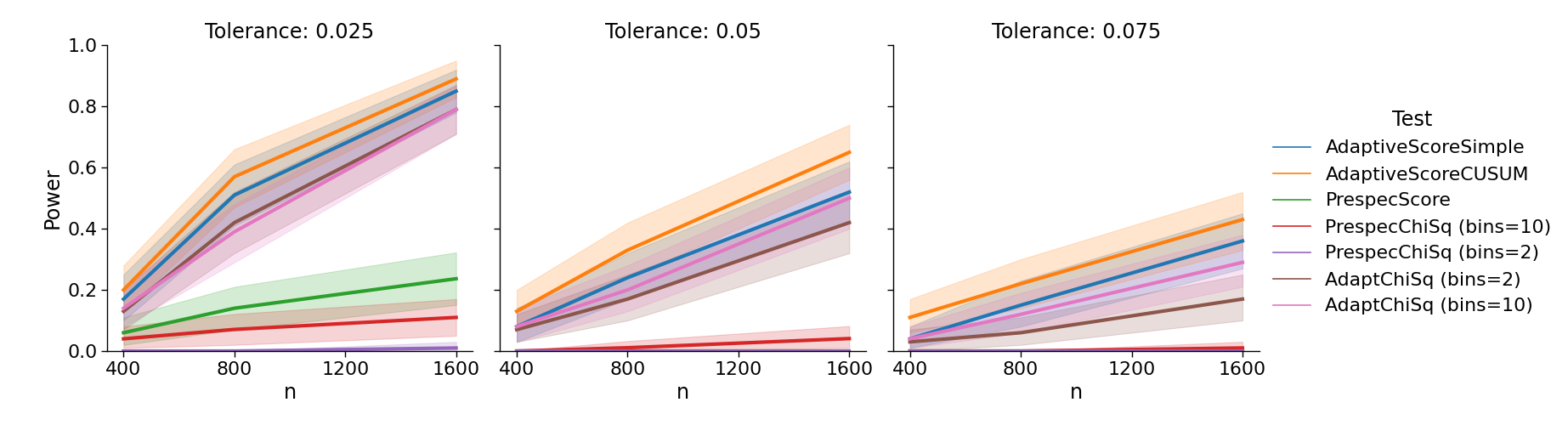}
 \includegraphics[width=0.28\textwidth]{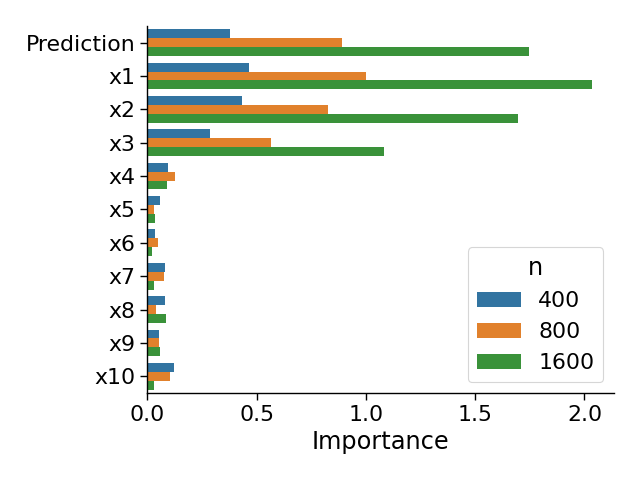}
	
	\caption{
		Testing for strong calibration of a misspecified logistic regression model (top) and a random forest model (bottom).
        Power is plotted against audit dataset sizes $n$ on the left, where 95\% confidence intervals are given by the shaded areas.
        Variable importance plots are on the right.
  Results are from 100 simulation replicates.
	}
 \vspace{-0.2in}
    \label{fig:power_mini}
\end{figure}

We test the two-sided null hypothesis \eqref{eq:hypo_test} for two algorithms: a logistic regression model (LR) that incorrectly assumes the logit is linear with respect to the variables and a RF.
A RF is not misspecified, but it must approximate the piecewise linear relationship and thus converges slowly to the true risk.
For tolerance levels $\delta=0.025,0.05$, and 0.075, the poorly calibrated subgroups had prevalences 0.60, 0.45, and 0.25 for the LR model and 0.8, 0.5, and 0.3 for the RF model, respectively.

\texttt{AdaptiveScoreCUSUM} consistently outperformed other methods across all settings (Figure~\ref{fig:power_mini}).
Tests that relied on a prespecified axis performed the worst.
Adaptive score-based tests demonstrated major improvements on the adaptive chi-squared test; in certain settings, power more than doubles.
This improvement is particularly evident at higher tolerance levels, where the poorly calibrated subgroup is smaller and it becomes even more important to extract much signal from the data as possible.
\texttt{AdaptiveScoreCUSUM} offered the biggest improvements over \texttt{AdaptiveScoreSimple} in smaller sample sizes, because the residual models can be quite inaccurate in such settings and, by restricting $\gamma$ to $0$, \texttt{AdaptiveScoreSimple} has difficulty isolating regions with poor calibration.

If we investigate the results from \texttt{AdaptiveScoreCUSUM} for the LR model, we find that the random forest-based detector maximized the test statistics in a majority of the cases.
This is unsurprising, given that the subgroup structure in  equation \eqref{eq:data_sim} matches the structure learned by recursive partitioning.
Moreover, the VI plots show that the detection models indeed recoreved the misspecified subgroup, as the variables $\hat{p}(x)$, and $x_1$, $x_2$ were assigned the highest importance.


Similarly, we investigate test results for the RF model.
As the audit dataset increased, the test statistic was more often maximized by kernel logistic regression.
This illustrates how RFs are not very powerful for detecting miscalibration of a RF, because we have already extracted most of the signal from the data using recursive partitioning.
Kernel logistic regression uses an entirely different approach, so it is better at extracting the remaining signal.
From the associated VI plots, we see that $x_1, x_2$, $\hat{p}(x)$, and $x_3$ are now the most important for detecting poor calibration.
This likely reflects the fact that kernel logistic regression converges faster than RF to the true probabilities in certain regions.

\section{Predicting one-year mortality using Electronic Health Record data}
\label{sec:zsfg}

We now compare the proposed procedure to existing methods in an analysis of Electronic Health Record (EHR) data from the Zuckerberg San Francisco General Hospital (ZSFG).
We trained RFs to predict one-year mortality using data from 2019 to 2022.
Input features to the RF included 31 demographic variables as well as ICD-10 diagnosis codes from the patient's record.
Across 40 replicates, we randomly assigned 250,000 patients to train the RF and randomly selected $n=4000,6000,8000$ observations for testing strong calibration.
We tested the two one-sided null hypotheses separately to differentiate between over- and under-estimation of the mortality rate.
For added interpretability, we tested for strong calibration with respect to only the demographic variables.

\begin{figure}
	\centering
        \vspace{-0.1in}
         \includegraphics[width=0.63\textwidth]{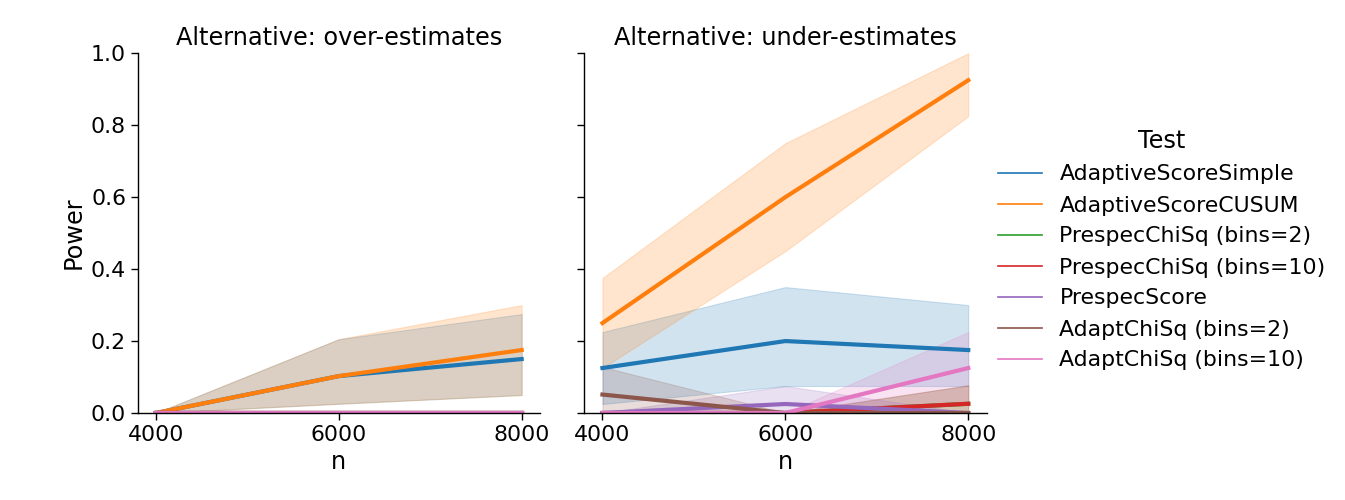}
         \includegraphics[width=0.23\textwidth]{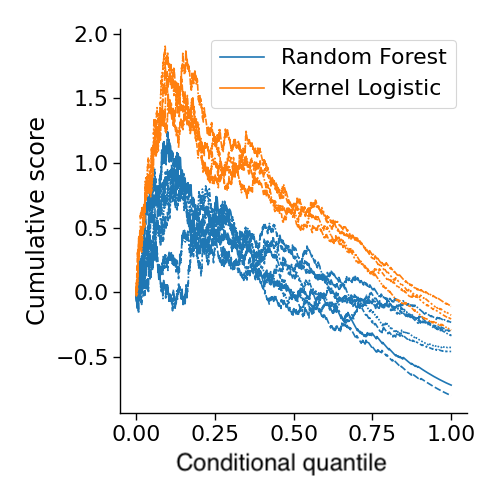}
        \includegraphics[width=0.36\textwidth]{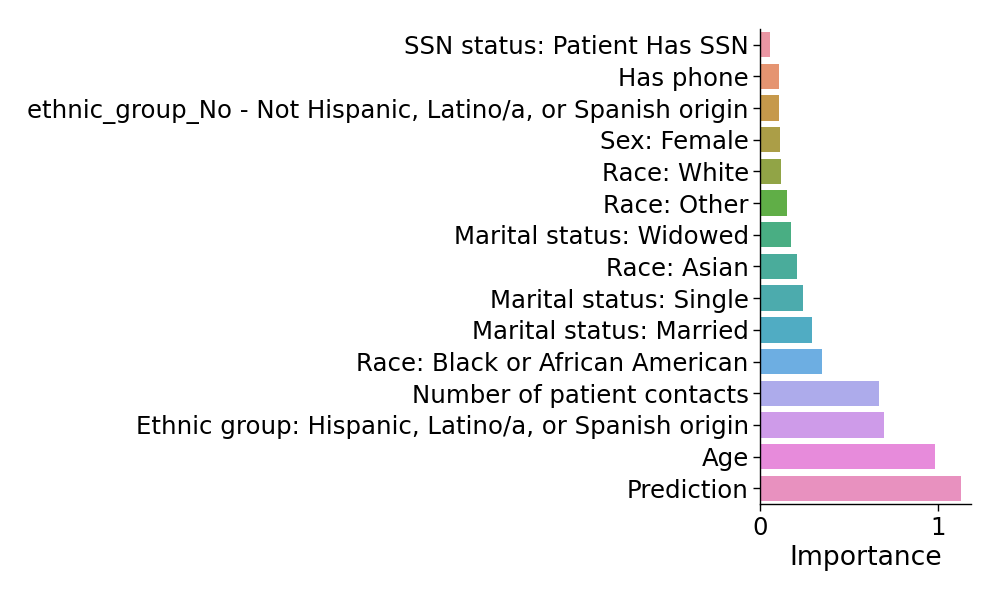}
        \includegraphics[width=0.6\textwidth]{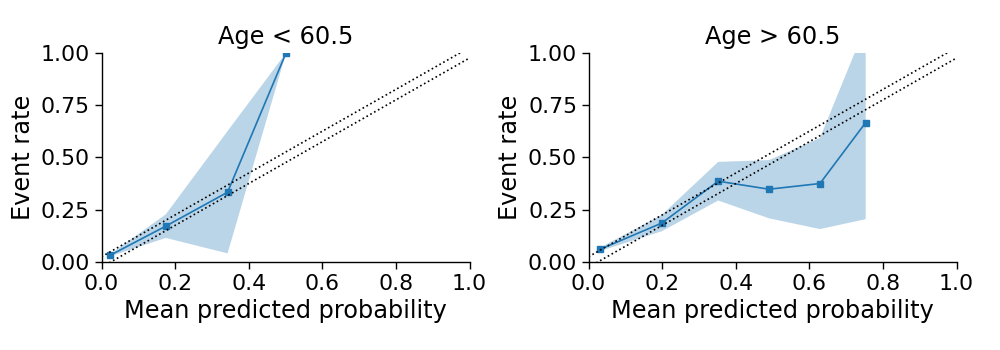}
        \caption{
        Testing if a risk prediction model for one-year mortality over- or under-estimates the true risk in some subgroup, beyond the tolerance level of $\delta = 0.025$.
        Top left: power across audit dataset sizes $n$.
        Top right: an example control chart plotting cumulative score.
        Bottom left: variable importance plot for the top 15 variables.
        Bottom right: calibration curves stratified by age, where the dotted lines correspond to calibration within tolerance $\delta$.
        }
        \vspace{-0.2in}
        \label{fig:zsfg}
\end{figure}

Figure~\ref{fig:zsfg} shows that there is strong evidence that the RF model under-estimates the true risks and a lack of evidence of over-estimation.
The rejection rate of \texttt{AdaptiveScoreCUSUM} reached 100\%, which was \textit{more than five times higher than that of other methods}.
Similar to that in our simulation study, we again find that kernel logistic regression was most often selected as the maximizer of the test statistic.
Analyzing the shape of the CUSUM plot from a sample replicate, we see that there is a sharp increase in the cumulative score process, and then a large drop.
This suggests that there is a small subgroup identified by the kernel logistic model that is very poorly calibrated.
VI plots further suggest that the most important variables that characterize this poorly calibrated group are age and the original risk prediction.
Based on these findings, we plotted the calibration curves stratified by age and indeed find that subjects with under-estimated risks are defined by the intersection of patients under 60 years old and with risk predictions over 40\%.
Diagnostic plots like these can help model developers better understand where their algorithm is unreliable, which can be used to restrict model usage to certain subgroups, inform model revision, and more.

\section{Discussion}

We have presented an adaptive score-based CUSUM procedure for testing if a given ML algorithm is poorly calibrated for some subgroup.
As shown in the empirical experiments, the procedure vastly outperforms existing methods.
The key insight of this method is that if one has an estimate of expected residuals at each observation, we can transform the problem of subgroup detection to one of changepoint detection.
Ordering observations by their predicted residuals, we should expect to see a change in the association between the observed and predicted residuals.
Using a changepoint formulation, we can fully leverage the natural ordering of the data and the information learned by the estimated residual models, and avoid unnecessary binning of the data.
Finally, we showed how accompanying control charts and VI plots can help model diagnostics and inform model revision.


\section*{Acknowledgments}
We thank Adarsh Subbaswamy and Karen Feng for helpful discussions and suggestions. We are grateful to Lucas Zier for sharing the ZSFG dataset.

This work was supported by the Food and Drug Administration (FDA) of the U.S. Department of Health and Human Services (HHS) as part of a financial assistance award Center  of  Excellence  in  Regulatory  Science  and  Innovation grant to University of California, San Francisco (UCSF) and Stanford University, U01FD005978 totaling \$79,250 with 100\% funded by FDA/HHS. The contents are those of the author(s) and do not necessarily represent the official views of, nor an endorsement, by FDA/HHS, or the U.S. Government.

\bibliographystyle{plainnat}
\bibliography{main}

\begin{thebibliography}{53}
\providecommand{\natexlab}[1]{#1}
\providecommand{\url}[1]{\texttt{#1}}
\expandafter\ifx\csname urlstyle\endcsname\relax
  \providecommand{\doi}[1]{doi: #1}\else
  \providecommand{\doi}{doi: \begingroup \urlstyle{rm}\Url}\fi

\bibitem[Barda et~al.(2021)Barda, Yona, Rothblum, Greenland, Leibowitz,
  Balicer, Bachmat, and Dagan]{Barda2021-qu}
Noam Barda, Gal Yona, Guy~N Rothblum, Philip Greenland, Morton Leibowitz, Ran
  Balicer, Eitan Bachmat, and Noa Dagan.
\newblock Addressing bias in prediction models by improving subpopulation
  calibration.
\newblock \emph{J. Am. Med. Inform. Assoc.}, 28\penalty0 (3):\penalty0
  549--558, March 2021.
\newblock URL \url{http://dx.doi.org/10.1093/jamia/ocaa283}.

\bibitem[Ben-Tal et~al.(2013)Ben-Tal, den Hertog, De~Waegenaere, Melenberg, and
  Rennen]{Ben-Tal}
Aharon Ben-Tal, Dick den Hertog, Anja De~Waegenaere, Bertrand Melenberg, and
  Gijs Rennen.
\newblock Robust solutions of optimization problems affected by uncertain
  probabilities.
\newblock \emph{Management Science}, 59\penalty0 (2):\penalty0 341--357, 2013.
\newblock \doi{10.1287/mnsc.1120.1641}.
\newblock URL \url{https://doi.org/10.1287/mnsc.1120.1641}.

\bibitem[Brown et~al.(1975)Brown, Durbin, and Evans]{Brown1975-sx}
R~L Brown, J~Durbin, and J~M Evans.
\newblock Techniques for testing the constancy of regression relationships over
  time.
\newblock \emph{J. R. Stat. Soc. Series B Stat. Methodol.}, 37\penalty0
  (2):\penalty0 149--192, 1975.
\newblock URL \url{http://www.jstor.org/stable/2984889}.

\bibitem[Buolamwini and Gebru(2018)]{Buolamwini2018-cb}
Joy Buolamwini and Timnit Gebru.
\newblock Gender shades: Intersectional accuracy disparities in commercial
  gender classification.
\newblock In Sorelle~A Friedler and Christo Wilson, editors, \emph{Proceedings
  of the 1st Conference on Fairness, Accountability and Transparency},
  volume~81 of \emph{Proceedings of Machine Learning Research}, pages 77--91,
  New York, NY, USA, 2018. PMLR.
\newblock URL \url{http://proceedings.mlr.press/v81/buolamwini18a.html}.

\bibitem[Chatterjee et~al.(2016)Chatterjee, Shi, and
  Garc{\'\i}a-Closas]{Chatterjee2016-lb}
Nilanjan Chatterjee, Jianxin Shi, and Montserrat Garc{\'\i}a-Closas.
\newblock Developing and evaluating polygenic risk prediction models for
  stratified disease prevention.
\newblock \emph{Nat. Rev. Genet.}, 17\penalty0 (7):\penalty0 392--406, July
  2016.
\newblock URL \url{http://dx.doi.org/10.1038/nrg.2016.27}.

\bibitem[Chung et~al.(2019)Chung, Kraska, Polyzotis, Tae, and
  Whang]{Chung2019-tz}
Yeounoh Chung, Tim Kraska, Neoklis Polyzotis, Ki~Hyun Tae, and Steven~Euijong
  Whang.
\newblock Slice finder: Automated data slicing for model validation.
\newblock In \emph{2019 {IEEE} 35th International Conference on Data
  Engineering ({ICDE})}, pages 1550--1553, April 2019.
\newblock URL \url{http://dx.doi.org/10.1109/ICDE.2019.00139}.

\bibitem[DiCiccio et~al.(2020)DiCiccio, Vasudevan, Basu, Kenthapadi, and
  Agarwal]{DiCiccio2020-yo}
Cyrus DiCiccio, Sriram Vasudevan, Kinjal Basu, Krishnaram Kenthapadi, and
  Deepak Agarwal.
\newblock Evaluating fairness using permutation tests.
\newblock In \emph{Proceedings of the 26th {ACM} {SIGKDD} International
  Conference on Knowledge Discovery \& Data Mining}, KDD '20, pages 1467--1477,
  New York, NY, USA, August 2020. Association for Computing Machinery.
\newblock URL \url{https://doi.org/10.1145/3394486.3403199}.

\bibitem[Duchi et~al.(2022)Duchi, Hashimoto, and Namkoong]{Duchi2022-kt}
John Duchi, Tatsunori Hashimoto, and Hongseok Namkoong.
\newblock Distributionally robust losses for latent covariate mixtures.
\newblock \emph{Oper. Res.}, September 2022.
\newblock URL \url{https://doi.org/10.1287/opre.2022.2363}.

\bibitem[Duchi and Namkoong(2021)]{Duchi2021-un}
John~C Duchi and Hongseok Namkoong.
\newblock Learning models with uniform performance via distributionally robust
  optimization.
\newblock \emph{The Annals of Statistics}, 49\penalty0 (3):\penalty0
  1378--1406, June 2021.
\newblock URL
  \url{https://projecteuclid.org/journals/annals-of-statistics/volume-49/issue-3/Learning-models-with-uniform-performance-via-distributionally-robust-optimization/10.1214/20-AOS2004.full}.

\bibitem[Dwork et~al.(2012)Dwork, Hardt, Pitassi, Reingold, and
  Zemel]{Dwork2012-aj}
Cynthia Dwork, Moritz Hardt, Toniann Pitassi, Omer Reingold, and Richard Zemel.
\newblock Fairness through awareness.
\newblock In \emph{Proceedings of the 3rd Innovations in Theoretical Computer
  Science Conference}, ITCS '12, pages 214--226, New York, NY, USA, January
  2012. Association for Computing Machinery.
\newblock URL \url{https://doi.org/10.1145/2090236.2090255}.

\bibitem[Eyuboglu et~al.(2022)Eyuboglu, Varma, Saab, Delbrouck, Lee-Messer,
  Dunnmon, Zou, and Re]{Eyuboglu2022-ho}
Sabri Eyuboglu, Maya Varma, Khaled~Kamal Saab, Jean-Benoit Delbrouck,
  Christopher Lee-Messer, Jared Dunnmon, James Zou, and Christopher Re.
\newblock Domino: Discovering systematic errors with {Cross-Modal} embeddings.
\newblock \emph{International Conference on Learning Representations}, May
  2022.
\newblock URL \url{https://openreview.net/forum?id=FPCMqjI0jXN}.

\bibitem[Feng et~al.(2022{\natexlab{a}})Feng, Gossmann, Pennello, Petrick,
  Sahiner, and Pirracchio]{Feng2022-az}
Jean Feng, Alexej Gossmann, Gene Pennello, Nicholas Petrick, Berkman Sahiner,
  and Romain Pirracchio.
\newblock Monitoring machine learning ({ML)-based} risk prediction algorithms
  in the presence of confounding medical interventions.
\newblock November 2022{\natexlab{a}}.
\newblock URL \url{http://arxiv.org/abs/2211.09781}.

\bibitem[Feng et~al.(2022{\natexlab{b}})Feng, Phillips, Malenica, Bishara,
  Hubbard, Celi, and Pirracchio]{Feng2022-mk}
Jean Feng, Rachael~V Phillips, Ivana Malenica, Andrew Bishara, Alan~E Hubbard,
  Leo~A Celi, and Romain Pirracchio.
\newblock Clinical artificial intelligence quality improvement: towards
  continual monitoring and updating of {AI} algorithms in healthcare.
\newblock \emph{npj Digital Medicine}, 5\penalty0 (1):\penalty0 1--9, May
  2022{\natexlab{b}}.
\newblock URL \url{https://www.nature.com/articles/s41746-022-00611-y}.

\bibitem[Foygel~Barber et~al.(2021)Foygel~Barber, Cand{\`e}s, Ramdas, and
  Tibshirani]{Foygel_Barber2021-gx}
Rina Foygel~Barber, Emmanuel~J Cand{\`e}s, Aaditya Ramdas, and Ryan~J
  Tibshirani.
\newblock The limits of distribution-free conditional predictive inference.
\newblock \emph{Inf Inference}, 10\penalty0 (2):\penalty0 455--482, June 2021.
\newblock URL
  \url{https://academic.oup.com/imaiai/article-pdf/10/2/455/38549621/iaaa017.pdf}.

\bibitem[Goff et~al.(2014)Goff, Lloyd-Jones, Bennett, Coady, D’Agostino,
  Gibbons, Greenland, Lackland, Levy, O’Donnell, Robinson, Schwartz, Shero,
  Smith, Sorlie, Stone, and Wilson]{Goff}
David~C. Goff, Donald~M. Lloyd-Jones, Glen Bennett, Sean Coady, Ralph~B.
  D’Agostino, Raymond Gibbons, Philip Greenland, Daniel~T. Lackland, Daniel
  Levy, Christopher~J. O’Donnell, Jennifer~G. Robinson, J.~Sanford Schwartz,
  Susan~T. Shero, Sidney~C. Smith, Paul Sorlie, Neil~J. Stone, and Peter W.~F.
  Wilson.
\newblock 2013 acc/aha guideline on the assessment of cardiovascular risk.
\newblock \emph{Circulation}, 129\penalty0 (25\_suppl\_2):\penalty0 S49--S73,
  2014.
\newblock \doi{10.1161/01.cir.0000437741.48606.98}.
\newblock URL
  \url{https://www.ahajournals.org/doi/abs/10.1161/01.cir.0000437741.48606.98}.

\bibitem[Gombay(2003)]{Gombay2003-ln}
Edit Gombay.
\newblock Sequential {Change-Point} detection and estimation.
\newblock \emph{Seq. Anal.}, 22\penalty0 (3):\penalty0 203--222, January 2003.
\newblock URL \url{https://doi.org/10.1081/SQA-120025028}.

\bibitem[Gombay(2017)]{Gombay2017-wb}
Edit Gombay.
\newblock Editor's special invited paper: On the efficient score vector in
  sequential monitoring.
\newblock \emph{Sequential Analysis}, 36\penalty0 (4):\penalty0 435--466,
  October 2017.
\newblock URL \url{https://doi.org/10.1080/07474946.2017.1394728}.

\bibitem[Guan(2023)]{Guan2023-is}
Leying Guan.
\newblock Localized conformal prediction: a generalized inference framework for
  conformal prediction.
\newblock \emph{Biometrika}, 110\penalty0 (1):\penalty0 33--50, February 2023.
\newblock URL
  \url{https://academic.oup.com/biomet/article-pdf/110/1/33/49160126/asac040.pdf}.

\bibitem[Hardt et~al.(2016)Hardt, Price, Price, and Srebro]{Hardt2016-fc}
Moritz Hardt, Eric Price, Eric Price, and Nati Srebro.
\newblock Equality of opportunity in supervised learning.
\newblock In D~D Lee, M~Sugiyama, U~V Luxburg, I~Guyon, and R~Garnett, editors,
  \emph{Advances in Neural Information Processing Systems 29}, pages
  3315--3323. Curran Associates, Inc., 2016.
\newblock URL
  \url{http://papers.nips.cc/paper/6374-equality-of-opportunity-in-supervised-learning.pdf}.

\bibitem[Hawkins(1991)]{Hawkins1991-cx}
Douglas~M Hawkins.
\newblock Diagnostics for use with regression recursive residuals.
\newblock \emph{Technometrics}, 33\penalty0 (2):\penalty0 221--234, 1991.
\newblock URL \url{http://www.jstor.org/stable/1269048}.

\bibitem[Hebert-Johnson et~al.(2018)Hebert-Johnson, Kim, Reingold, and
  Rothblum]{Hebert-Johnson2018-vb}
Ursula Hebert-Johnson, Michael Kim, Omer Reingold, and Guy Rothblum.
\newblock Multicalibration: Calibration for the
  ({{C}omputationally-Identifiable}) masses.
\newblock \emph{International Conference on Machine Learning}, 80:\penalty0
  1939--1948, 2018.
\newblock URL \url{https://proceedings.mlr.press/v80/hebert-johnson18a.html}.

\bibitem[Hosmer et~al.(1997)Hosmer, Hosmer, Le~Cessie, and
  Lemeshow]{Hosmer1997-ss}
D~W Hosmer, T~Hosmer, S~Le~Cessie, and S~Lemeshow.
\newblock A comparison of goodness-of-fit tests for the logistic regression
  model.
\newblock \emph{Stat. Med.}, 16\penalty0 (9):\penalty0 965--980, May 1997.
\newblock URL
  \url{http://dx.doi.org/10.1002/(sici)1097-0258(19970515)16:9<965::aid-sim509>3.0.co;2-o}.

\bibitem[Hosmer and Hjort(2002)]{Hosmer2002-mp}
David~W Hosmer and Nils~Lid Hjort.
\newblock Goodness-of-fit processes for logistic regression: simulation
  results.
\newblock \emph{Stat. Med.}, 21\penalty0 (18):\penalty0 2723--2738, September
  2002.
\newblock URL \url{http://dx.doi.org/10.1002/sim.1200}.

\bibitem[Hudson et~al.(2021)Hudson, Carone, and Shojaie]{Hudson2021-gm}
Aaron Hudson, Marco Carone, and Ali Shojaie.
\newblock Inference on function-valued parameters using a restricted score
  test.
\newblock May 2021.
\newblock URL \url{http://arxiv.org/abs/2105.06646}.

\bibitem[Ilvento(2020)]{Ilvento2020-bs}
Christina Ilvento.
\newblock Metric learning for individual fairness.
\newblock \emph{Symposium on Foundations of Responsible Computing}, 2020.
\newblock URL \url{http://arxiv.org/abs/1906.00250}.

\bibitem[Jankov{\'a} et~al.(2020)Jankov{\'a}, Shah, B{\"u}hlmann, and
  Samworth]{Jankova2020-kl}
Jana Jankov{\'a}, Rajen~D Shah, Peter B{\"u}hlmann, and Richard~J Samworth.
\newblock Goodness-of-fit testing in high dimensional generalized linear
  models.
\newblock \emph{J. R. Stat. Soc. Series B Stat. Methodol.}, 82\penalty0
  (3):\penalty0 773--795, July 2020.
\newblock URL \url{https://onlinelibrary.wiley.com/doi/10.1111/rssb.12371}.

\bibitem[Kim et~al.(2019)Kim, Ghorbani, and Zou]{Kim2019-bo}
Michael~P Kim, Amirata Ghorbani, and James Zou.
\newblock Multiaccuracy: {Black-Box} {Post-Processing} for fairness in
  classification.
\newblock In \emph{Proceedings of the 2019 {AAAI/ACM} Conference on {AI},
  Ethics, and Society}, AIES '19, pages 247--254, New York, NY, USA, January
  2019. Association for Computing Machinery.
\newblock URL \url{https://doi.org/10.1145/3306618.3314287}.

\bibitem[Lei and Wasserman(2014)]{Lei2014-uh}
Jing Lei and Larry Wasserman.
\newblock Distribution‐free prediction bands for non‐parametric regression.
\newblock \emph{Journal of the Royal Statistical Society, Series B},
  76\penalty0 (1), 2014.
\newblock URL
  \url{https://rss-onlinelibrary-wiley-com.offcampus.lib.washington.edu/doi/10.1111/rssb.12021}.

\bibitem[Lemeshow et~al.(2013)Lemeshow, Sturdivant, and
  Hosmer]{Lemeshow2013-mk}
Stanley Lemeshow, Rodney~X Sturdivant, and David~W Hosmer, Jr.
\newblock \emph{Applied Logistic Regression}.
\newblock Wiley \& Sons, Limited, John, 2013.
\newblock URL
  \url{https://openlibrary.org/books/OL38255075M/Applied_Logistic_Regression}.

\bibitem[Li et~al.(2021)Li, Namkoong, and Xia]{Li2021-bk}
Mike Li, Hongseok Namkoong, and Shangzhou Xia.
\newblock Evaluating model performance under worst-case subpopulations.
\newblock \emph{Conference on Neural Information Processing Systems}, 2021.
\newblock URL
  \url{https://proceedings.neurips.cc/paper_files/paper/2021/file/908075ea2c025c335f4865f7db427062-Paper.pdf}.

\bibitem[Lin et~al.(2002)Lin, Wei, and Ying]{Lin2002-gu}
D~Y Lin, L~J Wei, and Z~Ying.
\newblock Model-checking techniques based on cumulative residuals.
\newblock \emph{Biometrics}, 58\penalty0 (1):\penalty0 1--12, March 2002.
\newblock URL \url{http://dx.doi.org/10.1111/j.0006-341x.2002.00001.x}.

\bibitem[Lundberg and Lee(2017)]{Lundberg2017-xl}
Scott~M Lundberg and Su-In Lee.
\newblock A unified approach to interpreting model predictions.
\newblock \emph{Adv. Neural Inf. Process. Syst.}, pages 4765--4774, 2017.
\newblock URL
  \url{http://papers.nips.cc/paper/7062-a-unified-approach-to-interpreting-model-predictions.pdf}.

\bibitem[Luo et~al.(2022)Luo, Bhatnagar, Bai, Zhao, Wang, Xiong, Savarese,
  Ermon, Schmerling, and Pavone]{Luo2022-tg}
Rachel Luo, Aadyot Bhatnagar, Yu~Bai, Shengjia Zhao, Huan Wang, Caiming Xiong,
  Silvio Savarese, Stefano Ermon, Edward Schmerling, and Marco Pavone.
\newblock Local calibration: Metrics and recalibration.
\newblock \emph{Uncertain. Artif. Intell.}, 2022.
\newblock URL \url{https://openreview.net/pdf?id=BCg4lD8ice5}.

\bibitem[Maity et~al.(2021)Maity, Xue, Yurochkin, and Sun]{Maity2021-nm}
Subha Maity, Songkai Xue, Mikhail Yurochkin, and Yuekai Sun.
\newblock Statistical inference for individual fairness.
\newblock \emph{International Conference on Learning Representations}, March
  2021.
\newblock URL \url{http://arxiv.org/abs/2103.16714}.

\bibitem[Marx et~al.(2022)Marx, Zhao, Neiswanger, and Ermon]{Marx2022-jh}
Charles Marx, Shengjia Zhao, Willie Neiswanger, and Stefano Ermon.
\newblock Modular conformal calibration.
\newblock June 2022.
\newblock URL \url{https://proceedings.mlr.press/v162/marx22a/marx22a.pdf}.

\bibitem[Mitchell et~al.(2021)Mitchell, Potash, Barocas, D'Amour, and
  Lum]{Mitchell2021-wq}
Shira Mitchell, Eric Potash, Solon Barocas, Alexander D'Amour, and Kristian
  Lum.
\newblock Algorithmic fairness: Choices, assumptions, and definitions.
\newblock \emph{Annu. Rev. Stat. Appl.}, 8\penalty0 (1):\penalty0 141--163,
  March 2021.
\newblock URL \url{https://doi.org/10.1146/annurev-statistics-042720-125902}.

\bibitem[Montgomery(2013)]{Montgomery2013-sk}
Douglas~C Montgomery.
\newblock \emph{Statistical quality control}.
\newblock John Wiley \& Sons, Nashville, TN, 7 edition, 2013.

\bibitem[Platt(1999)]{Platt1999-lm}
John Platt.
\newblock Probabilistic outputs for support vector machines and comparisons to
  regularized likelihood methods.
\newblock \emph{Advances in large margin classifiers}, 10\penalty0
  (3):\penalty0 61--74, 1999.
\newblock URL
  \url{https://www.researchgate.net/file.PostFileLoader.html?id=540479d7d11b8bb1588b459d&assetKey=AS%3A273601008209920%401442242971560}.

\bibitem[Romano et~al.(2020)Romano, Barber, Sabatti, and
  Cand{\`e}s]{Romano2020-ua}
Yaniv Romano, Rina~Foygel Barber, Chiara Sabatti, and Emmanuel Cand{\`e}s.
\newblock With malice toward none: Assessing uncertainty via equalized
  coverage.
\newblock \emph{Harvard Data Science Review}, April 2020.
\newblock URL \url{https://hdsr.mitpress.mit.edu/pub/qedrwcz3/download/pdf}.

\bibitem[Ruoss et~al.(2020)Ruoss, Balunovi{\'c}, Fischer, and
  Vechev]{Ruoss2020-no}
Anian Ruoss, Mislav Balunovi{\'c}, Marc Fischer, and Martin Vechev.
\newblock Learning certified individually fair representations.
\newblock \emph{Conference on Neural Information Processing Systems}, February
  2020.
\newblock URL
  \url{https://proceedings.neurips.cc/paper/2020/file/55d491cf951b1b920900684d71419282-Paper.pdf}.

\bibitem[Subbaswamy et~al.(2021)Subbaswamy, Adams, and
  Saria]{Subbaswamy2021-np}
Adarsh Subbaswamy, Roy Adams, and Suchi Saria.
\newblock Evaluating model robustness and stability to dataset shift.
\newblock In Arindam Banerjee and Kenji Fukumizu, editors, \emph{Proceedings of
  The 24th International Conference on Artificial Intelligence and Statistics},
  volume 130 of \emph{Proceedings of Machine Learning Research}, pages
  2611--2619. PMLR, 2021.
\newblock URL \url{https://proceedings.mlr.press/v130/subbaswamy21a.html}.

\bibitem[Tsiatis(1980)]{Tsiatis1980-fp}
Anastasios~A Tsiatis.
\newblock A note on a goodness-of-fit test for the logistic regression model.
\newblock \emph{Biometrika}, 67\penalty0 (1):\penalty0 250--251, January 1980.
\newblock URL
  \url{https://academic.oup.com/biomet/article-pdf/67/1/250/6690321/67-1-250.pdf}.

\bibitem[Vaart(1998)]{vaart_1998}
A.~W. van~der Vaart.
\newblock \emph{Asymptotic Statistics}.
\newblock Cambridge Series in Statistical and Probabilistic Mathematics.
  Cambridge University Press, 1998.
\newblock \doi{10.1017/CBO9780511802256}.

\bibitem[Van~Calster and Vickers(2015)]{Van_Calster2015-ob}
Ben Van~Calster and Andrew~J Vickers.
\newblock Calibration of risk prediction models: impact on decision-analytic
  performance.
\newblock \emph{Med. Decis. Making}, 35\penalty0 (2):\penalty0 162--169,
  February 2015.
\newblock URL \url{http://dx.doi.org/10.1177/0272989X14547233}.

\bibitem[Van~Calster et~al.(2016)Van~Calster, Nieboer, Vergouwe, De~Cock,
  Pencina, and Steyerberg]{Van_Calster2016-ey}
Ben Van~Calster, Daan Nieboer, Yvonne Vergouwe, Bavo De~Cock, Michael~J
  Pencina, and Ewout~W Steyerberg.
\newblock A calibration hierarchy for risk models was defined: from utopia to
  empirical data.
\newblock \emph{J. Clin. Epidemiol.}, 74:\penalty0 167--176, June 2016.
\newblock URL \url{http://dx.doi.org/10.1016/j.jclinepi.2015.12.005}.

\bibitem[Vovk(2013)]{Vovk2013-ou}
Vladimir Vovk.
\newblock Conditional validity of inductive conformal predictors.
\newblock \emph{Mach. Learn.}, 92\penalty0 (2):\penalty0 349--376, September
  2013.
\newblock URL \url{https://doi.org/10.1007/s10994-013-5355-6}.

\bibitem[Vovk et~al.(2005)Vovk, Gammerman, and Shafer]{Vovk2005-wf}
Vladimir Vovk, Alexander Gammerman, and Glenn Shafer.
\newblock \emph{Algorithmic Learning in a Random World}.
\newblock Springer, Boston, MA, 2005.
\newblock URL \url{https://link.springer.com/book/10.1007/b106715}.

\bibitem[Vovk et~al.(2020)Vovk, Petej, Toccaceli, Gammerman, Ahlberg, and
  Carlsson]{Vovk2020-pi}
Vladimir Vovk, Ivan Petej, Paolo Toccaceli, Alexander Gammerman, Ernst Ahlberg,
  and Lars Carlsson.
\newblock Conformal calibrators.
\newblock \emph{Symposium on Conformal and Probabilistic Prediction and
  Applications}, 128:\penalty0 84--99, 2020.
\newblock URL \url{https://proceedings.mlr.press/v128/vovk20a.html}.

\bibitem[Wainwright(2019)]{Wainwright2019-sp}
Martin~J Wainwright.
\newblock \emph{{High-Dimensional} Statistics: A {Non-Asymptotic} Viewpoint}.
\newblock Cambridge University Press, February 2019.
\newblock URL
  \url{https://play.google.com/store/books/details?id=IluHDwAAQBAJ}.

\bibitem[Williamson and Feng(2020)]{Williamson2020-lx}
Brian~D Williamson and Jean Feng.
\newblock Efficient nonparametric statistical inference on population feature
  importance using shapley values.
\newblock \emph{International Conference on Machine Learning}, 2020.
\newblock URL
  \url{https://proceedings.icml.cc/static/paper_files/icml/2020/3042-Paper.pdf}.

\bibitem[Xue et~al.(2020)Xue, Yurochkin, and Sun]{Xue2020-nt}
Songkai Xue, Mikhail Yurochkin, and Yuekai Sun.
\newblock Auditing {ML} models for individual bias and unfairness.
\newblock \emph{International Conference on Artificial Intelligence and
  Statistics}, March 2020.
\newblock URL \url{http://proceedings.mlr.press/v108/xue20a/xue20a.pdf}.

\bibitem[Zhang et~al.(2021)Zhang, Ding, and Yang]{Zhang2021-oh}
Jiawei Zhang, Jie Ding, and Yuhong Yang.
\newblock Is a classification procedure good {Enough?---A} {Goodness-of-Fit}
  assessment tool for classification learning.
\newblock \emph{J. Am. Stat. Assoc.}, pages 1--11, September 2021.
\newblock URL \url{https://doi.org/10.1080/01621459.2021.1979010}.

\bibitem[Zhao et~al.(2020)Zhao, Ma, and Ermon]{Zhao2020-md}
Shengjia Zhao, Tengyu Ma, and Stefano Ermon.
\newblock Individual calibration with randomized forecasting.
\newblock In Hal~Daum{\'e} Iii and Aarti Singh, editors, \emph{Proceedings of
  the 37th International Conference on Machine Learning}, volume 119 of
  \emph{Proceedings of Machine Learning Research}, pages 11387--11397. PMLR,
  2020.
\newblock URL \url{https://proceedings.mlr.press/v119/zhao20e.html}.

\end{thebibliography}

\appendix

\section{Extension: Testing nonzero $\epsilon$}

To test for poorly calibrated subgroups with some minimum prevalence $\epsilon > 0$, the only modification needed is to constrain the set of detectors to those for which 
\begin{align}
	\mathbb{E}\left[\mathbbm{1}\{h(X) > 0 \} \right] > \epsilon.
	\label{eq:min_prev}
\end{align}
So we can use essentially the same sample-splitting or CV procedure, except we only consider thresholds $\gamma$ for which the corresponding detector satisfies \eqref{eq:min_prev}.

\section{Proofs}

Below we present proofs for all the theoretical results in the main manuscript.
We use $c_k$ to denote positive constants.

\paragraph{Proof of Theorem~\ref{theorem:super_unif}}
\begin{proof}
	It suffices to prove that conditional on the training data, the test statistic for the distribution with conditional probabilities equal to $\hat{p}_{\delta}$ stochastically dominates the test statistic for any other distribution under the null with conditional probabilities equal to $p_0$. 
	To do this, we use a coupling argument.
	
	Given any $x$, we can generate binary random variables (RVs) $Y$ and $\tilde{Y}$ where $\Pr(Y = 1|x) = p_0(x)$, $\Pr(\tilde{Y} = 1|x) = \hat{p}_{\delta}(x)$, and $Y \le \tilde{Y}$ as follows.
	First, sample a standard uniform random variable $U$. Then let $Y = \mathbbm{1}\{U \le p_0(x)\}$ and $\tilde{Y} = \mathbbm{1}\{U \le p_\delta(x)\}$.
	As such, the above conditions are satisfied.
	
	Using this procedure, we can generate coupled outcomes for observations in the test partition (i.e. $i = n_1 + 1,\cdots, n$).
	Consequently,  the test statistics on the coupled test data must satisfy
	$$
	\hat{T}_n^{(split)} = 
	\sup_{h \in \widehat{\mathcal{H}}_{+,\Lambda}}
	\frac{1}{n_2}
	\sum_{i=n_1 + 1}^{n}
	(Y_i - \hat{p}_{\delta}(X_i)) h(X_i)
	\le 
	\tilde{T}_n^{(split)} = 
	\sup_{h \in \widehat{\mathcal{H}}_{+,\Lambda}}
	\frac{1}{n_2}
	\sum_{i=n_1 + 1}^{n}
	(\tilde{Y}_i - \hat{p}_{\delta}(X_i)) h(X_i).
	$$
	As such, $\hat{T}_n^{(split)}$ stochastically dominates $\tilde{T}_n^{(split)}$.
\end{proof}

\paragraph{Proof for Theorem~\ref{theorem:power}}
\begin{proof}
	Below, we use the $\mathbbm{P}_{n_1 + 1:n}$  and $\mathbbm{P}$ to denote the empirical average over the test data split and the expectation, respectively.
	
	We begin with determining the minimum value of $\tau_{\alpha}$ to control Type I error.
	In particular, we must perform a multiplicity correction to account for the multiple residual models being tested.
	By a union bound, we have that
	\begin{align}
		& \Pr\left(
		\max_{h \in \widehat{H}_{+,\Lambda}}
		\left(\mathbbm{P}_{n_1 + 1: n} - \mathbbm{P}\right)
		(Y - \hat{p}_{\delta}(X)) h(X)
		> \tau_{\alpha}
		\right)\\
		\le & 
		\left |\Lambda \right|
		\max_{\lambda \in \Lambda}
		\Pr\left(
		\sup_{\gamma \ge 0}
		\left(\mathbbm{P}_{n_1 + 1: n} - \mathbbm{P}\right)
		(Y - \hat{p}_{\delta}(X)) \hat{g}_{\lambda,n}(X) \mathbbm{1}\left\{\hat{g}_{\lambda,n} \ge \gamma \right\}
		> \tau_{\alpha}
		\right).
		\label{eq:union}
	\end{align}
	Applying Theorem 4.10 in \citep{Wainwright2019-sp}, we have for any $\lambda\in \Lambda$ and $b\ge 0$ that
	\begin{align}
		\Pr\left(
		\sup_{\gamma \ge 0}
		\left(\mathbbm{P}_{n_1 + 1: n} - \mathbbm{P}\right) 
		(Y - \hat{p}_{\delta}(X)) \hat{g}_{\lambda,n}(X) \mathbbm{1}\left\{\hat{g}_{\lambda,n} \ge \gamma \right\}
		> 2 \mathcal{R}_{\lambda} + b
		\right)
		&\le \exp\left(-\frac{n_2b^2}{2c_1^2}\right)
		\label{eq:pr_bd}
	\end{align}
	where $\mathcal{R}_{\lambda}$ is an upper-bound of the Rademacher complexity for the function class
	$$
	\left\{
	x\mapsto \hat{h}_{\lambda, \gamma, n}(x) =  \hat{g}_{\lambda,n}(x) \mathbbm{1}\left\{\hat{g}_{\lambda,n}(x) \ge \gamma \right\}: \gamma \ge 0
	\right\}.
	$$
	Because the set of functions $\{x \mapsto \mathbbm{1}\{\hat{g}_{\lambda, n}(x) > \gamma\} : \gamma \ge 0\}$ has VC dimension 1, we have by an application of Lemma 4.14 in \citep{Wainwright2019-sp} that
	\begin{align}
		\mathcal{R}_{\lambda} \le c_2 \sqrt{\frac{\log(n_2+1)}{n_2}}.
		\label{eq:rademacher}
	\end{align}
	Plugging \eqref{eq:rademacher} and \eqref{eq:pr_bd} into \eqref{eq:union}, we find that by setting
	\begin{align}
		\tau_\alpha \ge c_3 \sqrt{\frac{\log(|\Lambda|(n_2+1)/\alpha)}{n_2}},
		\label{eq:tau_lower}
	\end{align}
	controls the finite-sample Type I error at level $\alpha$.
	
	Now suppose $n_1$ is chosen so that
	$$
	\mathbb{P}\left[\left(Y - \hat{p}_{\delta}(X)\right) h_{0,\gamma}(X) \right]
	- c_3 n_1^{-\omega/2}
	\ge \tau_{\alpha}.
	$$
	Note that by Cauchy Schwarz, we have that
	\begin{align*}
		\left|
		\mathbb{P} \left[
		\left((Y - \hat{p}_{\delta}(X)\right)
		\left (\hat{h}_{\lambda, \gamma, n}(X) -  h_{0,\gamma}(X)\right)
		\right]
		\right|
		& \le \ 
		c_4
		\sqrt{
			\mathbb{P}
			\left \|\hat{h}_{\lambda, \gamma, n}(X) -  h_{0,\gamma}(X)\right\|^2
		}.
	\end{align*}
	So conditional on the set $\mathcal{S}_{\lambda, \omega}^{(n_1)}$, the difference in the expected score is no greater than $c_5 n_1^{-\omega/2}$ for some $c_5 > 0$.
	Because $((Y - \hat{p}_{\delta}(X)) h_{0,\gamma}(X)$ is sub-gaussian, we have by Chernoff's bound that
	\begin{align*}
		& \Pr\left(\hat{T}_n^{(split)} < \tau_{\alpha}| \mathcal{S}_{\lambda, \omega}^{(n_1)}\right)\\
		\le &
		\Pr\left(
		(\mathbb{P}_{n_1 + 1:n} - \mathbb{P}) \left(Y - \hat{p}_{\delta}(X)\right) \hat{h}_{\lambda,\gamma,n}(X)
		+ \mathbb{P} \left(Y - \hat{p}_{\delta}(X)\right) \left(\hat{h}_{\lambda,\gamma,n}(X) - h_{0,\gamma}(X)\right)
		< \tau_\alpha - \mathbb{P} \left(Y - \hat{p}_{\delta}(X)\right) h_{0,\gamma}(X)
		\right)\\
		\le & \exp\left(-\frac{
			n_2\left(\mathbb{P} \left(Y - \hat{p}_{\delta}(X)\right) h_{0,\gamma}(X)
			- c n_1^{-\omega/2}
			- \tau_\alpha \right)^2
		}{c_6}\right).
	\end{align*}
	Plugging in \eqref{eq:tau_lower} to the above expression gives us our desired result.
\end{proof}

\paragraph{Proof of Theorem~\ref{theorem:cv}}
\begin{proof}
	We will use $\mathbb{P}_{n,k}$ to denote the empirical mean with respect to fold $k$ for $k = 1,\cdots, K$ and $\mathbb{P}_{n,-k}$ denote the mean with respect to data from all but the $k$th fold.
	Consider the decomposition
	\begin{align}
		\sqrt{n}\left(
		\begin{matrix}
			\mathbb{P}_{n,1}
			\left(Y - \hat{p}_{\delta}(X) \right) \hat{h}_{\lambda, \gamma,n}^{(-1)}(X)\\
			\vdots\\
			\mathbb{P}_{n,K}
			\left(Y - \hat{p}_{\delta}(X) \right) \hat{h}_{\lambda, \gamma,n}^{(-K)}(X)\\
		\end{matrix}
		\right)
		= & \
		\sqrt{n}\left(
		\begin{matrix}
			\mathbb{P}_{n,1}
			\left(Y - \hat{p}_{\delta}(X) \right) \bar{h}_{\lambda, \gamma,n}(X)\\
			\vdots\\
			\mathbb{P}_{n,K}
			\left(Y - \hat{p}_{\delta}(X) \right) \bar{h}_{\lambda, \gamma,n}(X)
		\end{matrix}
		\right)\\
		& \ + \sqrt{n}\left(
		\begin{matrix}
			\left(\mathbb{P}_{n,1} - \mathbb{P}\right)
			\left(Y - \hat{p}_{\delta}(X) \right)
			\left(
			\hat{h}_{\lambda, \gamma,n}^{(-1)}(X)
			- \bar{h}_{\lambda, \gamma,n}(X)
			\right)
			\\
			\vdots\\
			\left(\mathbb{P}_{n,K} - \mathbb{P}\right)
			\left(Y - \hat{p}_{\delta}(X) \right)
			\left(
			\hat{h}_{\lambda, \gamma,n}^{(-K)}(X)
			- \bar{h}_{\lambda, \gamma,n}(X)
			\right)
		\end{matrix}
		\right)
		\label{eq:remainder_emp}
		\\
		& \ + \sqrt{n}\left(
		\begin{matrix}
			\mathbb{P}
			\left(Y - \hat{p}_{\delta}(X) \right)
			\left(
			\hat{h}_{\lambda, \gamma,n}^{(-1)}(X)
			- \bar{h}_{\lambda, \gamma,n}(X)
			\right)
			\\
			\vdots\\
			\mathbb{P}
			\left(Y - \hat{p}_{\delta}(X) \right)
			\left(
			\hat{h}_{\lambda, \gamma,n}^{(-K)}(X)
			- \bar{h}_{\lambda, \gamma,n}(X)
			\right)
		\end{matrix}
		\right).
		\label{eq:remainder_exp}
	\end{align}
	
	First, we show that \eqref{eq:remainder_exp} is equal to zero.
	To see this, we have by the law of iterated expectations that
	\begin{align}
		\mathbb{P}
		\left(Y - \hat{p}_{\delta}(X) \right)
		\left(
		\hat{h}_{\lambda, \gamma,n}^{(-k)}(X)
		- \bar{h}_{\lambda, \gamma,n}(X)
		\right)
		& =
		\mathbb{P}_{n,-k}
		\left[
		\mathbb{P}
		\left[
		Y - \hat{p}_{\delta}(X)\mid X
		\right]
		\left(
		\hat{h}_{\lambda, \gamma,n}^{(-k)}(X)
		- \bar{h}_{\lambda, \gamma,n}(X)
		\right)
		\right],
		\label{eq:exp_exp}
	\end{align}
	where we marginalize over data in the $k$th fold and then marginalize over all but the $k$th fold.
	By the definition of $\bar{h}_{\lambda, \gamma,n}$, we have that the right hand side of \eqref{eq:exp_exp} is equal to zero.
	
	Next, we show that \eqref{eq:remainder_emp} is $o_p(1)$.
	Because the class of detectors varies across $n$, we will apply Theorem 19.28 in \citep{vaart_1998}, which is a generalization of Donsker's theorem that allows the indexing class to vary over $n$.
	To apply this result, note that the Lindeberg condition is satisfied, as residuals and detectors in the set $\widehat{H}_{\Lambda}$ are bounded.
	In addition, the bracketing entropy requirements are also satisfied.
	Thus we have that the stochastic process
	\begin{align}
		\left\{
		(\lambda, \gamma)
		\mapsto
		\sqrt{n}
		\left(\mathbb{P}_{n,k} - \mathbb{P}\right)
		\left(Y - \hat{p}_{\delta}(X) \right)
		\left(
		\hat{h}_{\lambda, \gamma,n}^{(-k)}(X)
		- \bar{h}_{\lambda, \gamma,n}(X)
		\right)
		\right\}
	\end{align}
	converges to a mean-zero Gaussian process with covariance function $\Sigma((\lambda_1, \gamma_1), (\lambda_2, \gamma_2))$ equal to
	\begin{align}
		\lim_{n}
		\Cov\left(
		\left(Y - \hat{p}_{\delta}(X) \right)
		\left(
		\hat{h}_{\lambda_1, \gamma_1,n}^{(-k)}(X)
		- \bar{h}_{\lambda_1, \gamma_1,n}(X)
		\right)
		,
		\left(Y - \hat{p}_{\delta}(X) \right)
		\left(
		\hat{h}_{\lambda_2, \gamma_2,n}^{(-k)}(X)
		- \bar{h}_{\lambda_2, \gamma_2,n}(X)
		\right)
		\right).
		\label{eq:cov_lim}
	\end{align}
	By Assumption~\ref{assump:converges}, \eqref{eq:cov_lim} converges to zero.
	Thus we have that
	\begin{align}
		\sqrt{n}\left(
		\begin{matrix}
			\left(\mathbb{P}_{n,1} - \mathbb{P}\right)
			\left(Y - \hat{p}_{\delta}(X) \right)
			\left(
			\hat{h}_{\lambda, \gamma,n}^{(-k)}(X)
			- \bar{h}_{\lambda, \gamma,n}(X)
			\right)
			\\
			\vdots\\
			\left(\mathbb{P}_{n,K} - \mathbb{P}\right)
			\left(Y - \hat{p}_{\delta}(X) \right)
			\left(
			\hat{h}_{\lambda, \gamma,n}^{(-k)}(X)
			- \bar{h}_{\lambda, \gamma,n}(X)
			\right)
		\end{matrix}
		\right)
		=o_p(1)
	\end{align}
	
	Combining the above results, we have established that
	\begin{align}
		\sup_{\lambda, \gamma}
		\sqrt{n}
		\left(
		\sum_{k=1}^K
		\mathbb{P}_{n,k}
		\left(Y - \hat{p}_{\delta}(X) \right) \hat{h}_{\lambda, \gamma,n}^{(-k)}(X)
		-
		\sum_{k=1}^K
		\mathbb{P}_{n,k}
		\left(Y - \hat{p}_{\delta}(X) \right) \bar{h}_{\lambda, \gamma,n}(X)
		\right)
		= o_p(1).
		\label{eq:general_equiv}
	\end{align}
	
	Having established that the remainder terms \eqref{eq:remainder_emp} and \eqref{eq:remainder_exp} are negligible, we can use the same arguments used to prove Theorem~\ref{theorem:super_unif} to prove that the setting the critical value $\tau_{\alpha}$ to the $1-\alpha$ quantile of
	\begin{align}
		T_{n,>}^{*(CV, oracle)}
		:= \sum_{k=1}^K
		\sum_{(X_i, Y_i^*) \in V_k}
		\left(Y^*_i - \hat{f}_{\delta}(X_i) \right) \bar{h}_{\lambda, \gamma,n}(X_i),
	\end{align}
	where $Y_i^*$ is a resampled binary RV that is equal to one with probability $\hat{p}_{\delta}(X_i)$, controls the Type I error asymptotically.

	In practice, $\bar{h}_{\lambda, \gamma,n}$ is unknown.
	So instead, we calculate the quantile for $T_{n,>}^{*(CV)}$, which plugs in the estimated detectors instead.	
	To prove that this plug-in approach maintains asymptotic Type I error control, we must show that the difference between $T_{n,>}^{*(CV)}$ and $T_{n,>}^{*(CV, oracle)}$ is asymptotically negligible.
	Let $\mathbb{P}_{n,k}^*$ denote the empirical mean in the $k$-th fold with respect to the resampled outcomes $(Y^*_1,\cdots, Y^*_n)$.
	Similarly, let $\mathbb{P}^*$ denote the expectation with respect to the distribution with conditional probability equal to $\hat{p}_{\delta}$.
	Consider the following decomposition for each $k = 1,\cdots, K$:
	\begin{align}
		\sqrt{n}
		\mathbb{P}_{n,k}^*
		\left(Y^* - \hat{f}_{\delta}(X) \right) 
		\left(
		\hat{h}_{\lambda, \gamma,n}^{(-k)}(X)
		- \bar{h}_{\lambda, \gamma,n}(X)
		\right)
		= & \
		\sqrt{n}
		\left(\mathbb{P}_{n,k}^* - \mathbb{P}^* \right)
		\left(Y^* - \hat{p}_{\delta}(X) \right)
		\left(\bar{h}_{\lambda, \gamma,n}^{(-k)}(X) - \bar{h}_{\lambda, \gamma,n}(X)\right)
		\label{eq:boot_rem1}
		\\
		& + \
		\sqrt{n}
		\mathbb{P}^*
		\left(Y - \hat{p}_{\delta}(X) \right)
		\left(\bar{h}_{\lambda, \gamma,n}^{(-k)}(X) - \bar{h}_{\lambda, \gamma,n}(X)\right).
		\label{eq:boot_rem2}
	\end{align}
	Using the same arguments as above, we have that \eqref{eq:boot_rem1} is $o_p(1)$ by Theorem 19.28 in \citep{vaart_1998} and \eqref{eq:boot_rem2} is equal to zero.
	Summing these results over all $k$, we have established that  the scaled difference between the calculated and oracle test statistic  $\frac{1}{\sqrt{n}} \left(T_{n,>}^{*(CV)} - T_{n,>}^{*(CV, oracle)}\right)$ is $o_p(1)$.
	Therefore, by Slutsky's theorem, the $1-\alpha$ quantile for $T_n^{*,(CV)}$ controls the Type I error at the desired rate.
\end{proof}

\paragraph{Proof of Theorem~\ref{theorem:super_unif_two_side}}
\begin{proof}
	We again use a coupling argument to prove that the modified statistic stochastically dominates the test statistic under any distribution ${p}_{0}$ satisfying the null hypothesis.
	In particular, consider the sampling procedure where $U$ is a standard uniform RV,
	$Y = \mathbbm{1}\{U \le {p}_{0}(X) \}$, $Y^{(-1)} = \mathbbm{1}\{U \le \hat{p}_{-\delta}(X) \}$, and $Y^{(1)} = \mathbbm{1}\{U \le \hat{p}_{\delta}(X) \}$.
	Thus the conditional probabilities $\Pr(Y = 1|X)$, $\Pr(Y^{(-1)} = 1|X)$, and $\Pr(Y^{(1)} = 1|X)$ are $p_0(X)$, $\hat{p}_{-\delta}(X)$, and $\hat{p}_{\delta}(X)$, respectively.
	Also, for any $h$ and $i$, we have that
	\begin{align}
		\left(Y - \hat{p}_{\delta \sign(\hat{h}_{\lambda, \gamma, n})}(X)\right)
		\hat{h}_{\lambda, \gamma, n}(X)
		\le
		\left(Y^{(\sign(\hat{h}_{\lambda, \gamma, n}))} - \hat{p}_{\delta \sign(\hat{h}_{\lambda, \gamma, n})}(X)\right)
		\hat{h}_{\lambda, \gamma, n}(X).
	\end{align}
	So if we use this procedure to resample the binary outcomes for the test partition, we would have that
	\begin{align*}
	&	\max_{\lambda \in \Lambda, \gamma\ge 0}
		\sum_{i=n_1+1}^n
		\left(Y_i - \hat{p}_{\delta \sign(\hat{h}_{\lambda, \gamma, n})}(X_i)\right)
		\hat{h}_{\lambda, \gamma, n}(X_i)\\
		\le&
		\max_{\lambda \in \Lambda, \gamma\ge 0}
		\sum_{i=n_1+1}^n
		\left(Y^{(\sign(\hat{h}_{\lambda, \gamma, n}))} - \hat{p}_{\delta \sign(\hat{h}_{\lambda, \gamma, n})}(X_i)\right)
		\hat{h}_{\lambda, \gamma, n}(X_i).
	\end{align*}
	This implies our desired result.
\end{proof}

\section{Additional simulation details}

For the residual models, we fit random forests and kernel logistic regression using the scikit-learn package.
We tuned the following hyperparameters for RF: maximum number of features $p=5$ versus $p=10$ and max depth of 4 versus 8.
For kernel logistic regression, we used an approximation of the polynomial kernel with degree two and subsequently fit a ridge-penalized logistic regression model, where we considered a regularization factor of $C = 1000, 100$, and $10$.

\section{Simulation study of Type I error control}
The goal of this simulation is to analyze the Type I error of the score-test in finite samples.
We test for strong calibration with tolerance $\delta = 0.025$.
For the original ML algorithm, we train a logistic regression model using 10,000 observations generated with the conditional log odds as
$$
\logit(p_{\text{orig}}(X)) = 0.6x_1 + 0.4 x_2 + 0.2 x_3.
$$
For the audit data, we simulated outcomes where the conditional probabilities were $p_{\text{orig}} + 0.025$ to maximize Type I error.
This simulation also reflects situations where ML algorithm is developed for one context and deployed in another, and one is interested in knowing if the ML algorithm is miscalibrated in some subgroup in the new target population.

We perform a one-sided test to determine if the predicted risks are under-estimates and a two-sided test.
We implement the CV-based testing procedures for which we have only established asymptotic control of the Type I error rate.
We do not include results from the single sample split, since we proved that it provides finite sample control of the Type I error rate.
The critical values were set to target a Type I error rate of 0.1.
Recall that the one-sided test calculates the critical value by sampling from the single worst-case null distribution, whereas the two-sided test relies on sampling upper bounds of the test statistic.
As such, we expect the observed Type I error rate to be lower (i.e. more conservative) for the two-sided test than the one-sided test.

\begin{figure}
	\centering
	\includegraphics[width=0.8\textwidth]{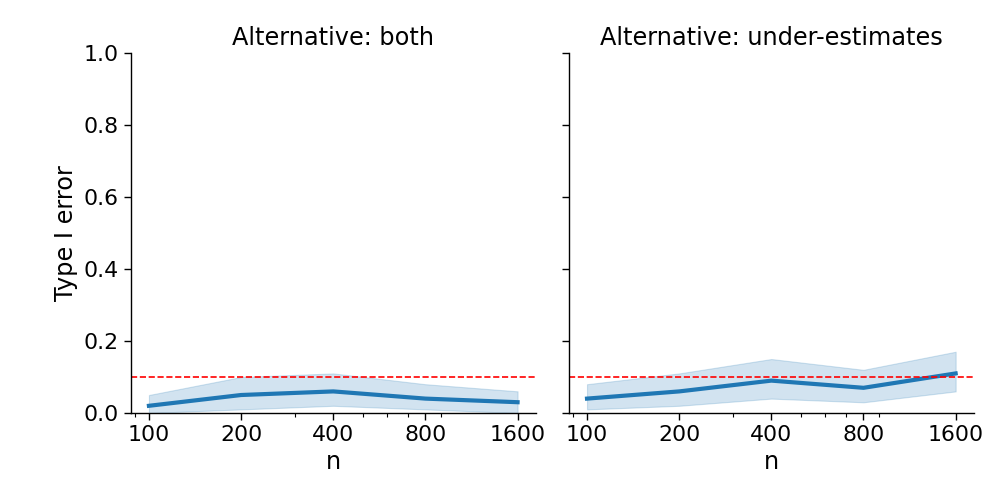}
	\caption{Simulation study of Type I error control. Note that the x--axis is shown on the log scale.}
	\label{fig:null}
\end{figure}

As shown in Figure~\ref{fig:null}, we observe that Type I error is controlled across a variety of audit dataset sizes, including $n = 100$.
So, even though our procedure only guarantees finite sample error rate control for the sample-splitting version, we are able to maintain Type I error control for small sample sizes even in the CV version because the assumptions needed for the CV procedure are very weak.

\end{document}